\documentclass[wcp]{jmlr}

\usepackage{longtable}

\usepackage{booktabs}
\usepackage{microtype}
\usepackage{hyperref}
\usepackage{mathtools}
\usepackage[capitalize,noabbrev]{cleveref}
\usepackage[textsize=tiny]{todonotes}
\usepackage{amsfonts}       
\usepackage{nicefrac}       
\usepackage{xcolor}         
\usepackage{subcaption}
\usepackage{wrapfig}
\usepackage{float}
\usepackage{tablefootnote}
\usepackage{lineno}

\DeclareMathOperator*{\argmin}{arg\,min}

\makeatletter
\let\Ginclude@graphics\@org@Ginclude@graphics 
\makeatother

\jmlrvolume{260}
\jmlryear{2024}
\jmlrworkshop{ACML 2024}
\title[Toward Data Efficient Model Merging between Different Datasets]{Toward Data Efficient Model Merging between Different Datasets without Performance Degradation}

\author{%
  \Name{Masanori Yamda}\Email{masanori.yamada@ntt.com} \\
  \addr NTT Social Informatics Laboratories
  \AND
  \Name{Tomoya Yamashita} \\
  \addr NTT Social Informatics Laboratories
  \AND
  \Name{Shin'ya Yamaguchi} \\
  \addr NTT Computer and Data Science Laboratories
  \AND
  \Name{Daiki Chijiwa} \\
  \addr NTT Computer and Data Science Laboratories
}

\editors{Vu Nguyen and Hsuan-Tien Lin}

\begin{document}

\maketitle

\begin{abstract}
Model merging is attracting attention as a novel method for creating a new model by combining the weights of different trained models. While previous studies reported that model merging works well for models trained on a single dataset with different random seeds, model merging between different datasets remains unsolved. In this paper, we attempt to reveal the difficulty in merging such models trained on different datasets and alleviate it. Our empirical analyses show that, in contrast to the single-dataset scenarios, dataset information needs to be accessed to achieve high accuracy when merging models trained on different datasets. However, the requirement to use full datasets not only incurs significant computational costs but also becomes a major limitation when integrating models developed and shared by others. 
To address this, we demonstrate that dataset reduction techniques, such as coreset selection and dataset condensation, effectively reduce the data requirement for model merging. 
In our experiments with SPLIT-CIFAR10 model merging, the accuracy is significantly improved by $31\%$ when using the full dataset and $24\%$ when using the sampled subset compared with not using the dataset.
\end{abstract}

\begin{keywords}
Model Merging, Mode Connectivity
\end{keywords}

\section{Introduction}
Model merging~\citep{leontev2020non,singh2020model,ainsworth2023git} is an approach to creating a new model that combines the weights contained in different deep neural networks~(DNNs). One scenario involves merging fine-tuned models from the same pre-trained model~\citep{goddard2024arcee,daheim2024model,leontev2020non,matena2022merging}. However, this paper focuses on the more general scenario of merging models trained with different random seeds, specifically in the context of image classification. This scenario is more challenging due to the differences in initialization. The recent discovery of a DNN property called Linear Mode Connectivity~(LMC)~\citep{frankle2020linear} brings us closer to successful model merging. The LMC indicates that if the optimal weights of two DNNs are given, they are trapped in the same basin of the loss landscape. This means that the model created from the average of these weights has a high accuracy since there is no loss barrier between the weights with LMC. Rahim~\citep{entezari2021role} and Ainsworth~\citep{ainsworth2023git} reported that two networks trained with different random seeds have an LMC if the neurons are properly aligned. 

Previous studies~\citep{ainsworth2023git,leontev2020non,singh2020model} reported that model merging works well for models trained on a single dataset with different random seeds, called model merging on a single dataset. On the other hand, model merging is difficult for models trained on different datasets, called model merging between different datasets. Merging knowledge from different datasets has practical significance. Thus, this study addresses the following research question.
\vspace{-0.5em}
\begin{itemize}
    \item RQ1: What makes model merging between different datasets more difficult than model merging on a single dataset?
\end{itemize}
\vspace{-0.5em}
To address RQ1, we focus on the difference between loss landscapes of the datasets that are incrementally increasing the gap from the original one. Specifically, we design a series of synthetic datasets based on MNIST, on which we analyze how the optimal basin in loss landscape differs due to the increasing gap. Previous work~\citep{ainsworth2023git} observed that for a single dataset, sufficiently close optimal weights reside within the same basin, effectively eliminating the loss barrier. In contrast, we find that in the case of different datasets, even if the optimal weights are close between these datasets, the loss barriers still remain, as shown in Fig.~\ref{fig:losslandscape-rmnist-90}. Moreover, we observed that overly close optimal weights yield a rather high loss barrier between them as shown in Fig.~\ref{fig:population}.

In this paper, we attempt to effectively merge models trained with different datasets by overcoming the challenges identified in our RQ1 analysis. Existing methods~\citep{entezari2021role,ainsworth2023git} remove the loss barrier by aligning neurons and averaging weights to merge models. The methods for aligning neurons can be categorized into (i) merging with weights~(MW) that aligns neurons based on their weights, and (ii) merging with weights and datasets~(MWD) that aligns neurons based on their weights and dataset information. Since there was no difference in accuracy between the two methods and methods using only weights are less costly, existing studies~\citep{ainsworth2023git,jordan2022repair} mainly focus on MW. However, in the model merging between the different datasets, it is not obvious whether the merging methods using only weights are sufficient, since the closest optimal weights have a large loss value on the datasets of each other as discussed in the previous paragraph. Thus, the following research question naturally arises.
\vspace{-0.5em}
\begin{itemize}
    \item RQ2: For model merging between different datasets, does the method of using weight and dataset information improve accuracy compared to using only weights?
\end{itemize}
\vspace{-0.5em}
In all cases in our experiments, MWD greatly improved accuracy compared to MW in contrast to the single dataset cases. Furthermore, to investigate dataset information requirements while minimizing reliance on optimization methods, such as the linear matching and Sinkhorn algorithm, we generate various alignments without a dataset. As a result, we found that no criteria used in the MW methods are positively correlated with the loss on the mixed dataset composed of the different datasets. On the other hand, the alignment criteria created from the dataset correlate well with the loss on the mixed dataset. Therefore, we conclude that dataset information is currently required for effective model merging between different datasets.

While model merging between different datasets can achieve high accuracy through neuron alignment with datasets, the necessity of utilizing full datasets presents a significant limitation. This limitation is particularly pronounced in scenarios when only the models are published without their datasets. Furthermore, the requirement of full datasets for model merging leads to significantly increased storage costs. In collaborative projects across multiple organizations, these limitations pose notable challenges. Sharing individually trained models is generally feasible, but exchanging large datasets often faces practical barriers, such as privacy, security, or logistical issues. Thus, we further investigate a third research question addressing these challenges.
\vspace{-0.5em}
\begin{itemize}
    \item RQ3: Can we reduce the data points required for effective model merging?
\end{itemize}
\vspace{-0.5em}
To address RQ3, we propose a method for model merging using a small surrogate dataset instead of a full-scale real dataset. We explore two approaches for generating a surrogate dataset: (i) Coreset selection~\citep{mirzasoleiman2020coresets,guo2022deepcore}, which aims to select a subset of the most informative training samples, and (ii) Dataset condensation~\citep{zhao2021dataset,nguyen2021dataset}, which distills the knowledge of a dataset into small numbers of data points to achieve high accuracy with less data. Our experiments demonstrate that model merging using both our proposed methods achieves higher accuracy than the baselines, which include the MW method and existing techniques for model merging between different datasets. In particular, it is surprising that high accuracy is achieved even when using a simple coreset selection with random sampling from each class. This is because the proposed method has the advantage of using the optimal weight information of each dataset. The neuron alignment operation keeps a low loss value on the mixed dataset, since its alignment strictly keeps the optimized loss value on the individual datasets. Our main contributions are summarized as follows:
\begin{itemize}
    \item We show that model merging becomes more difficult as datasets become more different and reveal that the difficulty of model merging is caused by the closest optimal weights having a large loss value on the datasets of each other as datasets become more different.
    \item We find that dataset information is currently required for effectively merging models between different datasets and that the alignment criteria created from the weights are not positively correlated with the performance.   
    \item We propose a method of merging models by using a surrogate dataset including random sampling instead of the real dataset. This approach greatly reduces the number of data points required. 
\end{itemize}
\vspace{-1.0em}
\section{Preliminary}
We present a short preliminary for model merging. First, we formulate the problem of model merging extended between different datasets. Then, we introduce permutation symmetry of neurons in DNN, which plays an important role in model merging. Finally, we introduce existing methods for model merging.
\vspace{-0.5em}
\subsection{Goal of Model Merging}
Let us consider merging the model parameters $\mathbf{w_{A}}$ and $\mathbf{w_{B}}$ trained on datasets generated from dataset probability distribution $P_A$ and $P_B$ respectively. The mixed dataset probability distribution\footnote{The mixed distribution refers to a combination of two separate datasets rather than the averaging of the individual data points as in mixup.} is defined as $P_{AB}\left(\mathbf{x},y,\alpha\right)=\left(1-\alpha\right) P_{A}\left(\mathbf{x},y\right)+\alpha P_{B}\left(\mathbf{x},y\right)$, where $\alpha\in\left[0,1\right]$ is a constant for the mixing ratio of datasets, for simplicity $\alpha=\frac{1}{2}$ is used. Note that the general $\alpha$ can be discussed in a similar way. The goal of model merging is to obtain operator $\star$ as follows
\begin{align}
\mathcal{L}_{AB}\left(\mathbf{w_{AB}}\right)\simeq\mathcal{L}_{AB}\left(\mathbf{w_{A}}\star\mathbf{w_{B}}\right)\label{eq:merge},
\end{align} where 
$\mathbf{w_{A}}=\argmin_{\mathbf{w}}\mathcal{L}_{A}\left(\mathbf{w}\right)$, $\mathbf{w_{B}}=\argmin_{\mathbf{w}}\mathcal{L}_{B}\left(\mathbf{w}\right)$, and
$\mathbf{w_{AB}}=\argmin_{\mathbf{w}}\mathcal{L}_{AB}\left(\mathbf{w}\right)$.
$\mathcal{L}_{A}\left(\mathbf{w}\right)$ denotes loss on $P_{A}$ as $\mathcal{L}_{A}=E_{P_{A}\left(\mathbf{x},y\right)}\left[-\log p\left(y|\mathbf{x},\mathbf{w}\right)\right]$. In this paper we assume for simplicity that Models A and B have the same architecture.

\subsection{Permutation Symmetry}
DNNs have neuron permutation symmetry, which is realized by rearranging the weights. Neuron permutation symmetry refers to the property that the output remains invariant with respect to the rearrangement of neurons. Let us consider a specific example of a simple feedforward network. Even if we apply a permutation matrix $\pi_{\ell}$ to the weights of the $\ell$-th layer $\mathbf{w_{\ell}}$, the output remains invariant if we apply the inverse of $\pi_{\ell}$ in the weights of the $(\ell+1)$-th layer. In other words, $\mathbf{w_{\ell+1}}\pi_{\ell}^{-1}\sigma\left(\pi_{\ell}\mathbf{w_{\ell}}\mathbf{x_{\ell}}\right)=\mathbf{w_{\ell+1}}\sigma\left(\mathbf{w_{\ell}}\mathbf{x_{\ell}}\right)$. Here, $\sigma$ represents the activation function, and $\mathbf{x_{\ell}}$ denotes the input of the $\ell$-th layer. To maintain invariant input-output relations in the neural network, we fix the input and output layers and only permute the neurons in the hidden layers.
By utilizing this freedom of rearrangement, we can permute the weights without changing the loss value.

\subsection{Model Merging}
Existing research~\citep{entezari2021role,ainsworth2023git} demonstrates that suitably aligning and averaging the weights to merge models can effectively minimize the loss on a single dataset. The suitable alignment and averaging of the weights are formulated as
\begin{align}
    \mathbf{w_{A}}\star\mathbf{w_{B}}=\left(1-\lambda\right)\mathbf{w_{A}}+\lambda\pi\left(\mathbf{w_{B}}\right),\label{eq:star-op}
\end{align}
where $\pi$ is permutation weights for all layers and $\lambda\in\left[0,1\right]$ is constant. We present two methods introduced in Ainsworth~\citep{ainsworth2023git}.

\noindent{\bf{WM:}} Weight Matching (WM) permutes the weights to reduce the L2 distance between weights of trained models as
\begin{align}
\argmin_{\pi}\left\Vert {\rm vec}\left(\mathbf{w_{A}}\right)-{\rm vec}\left(\pi\left(\mathbf{w_{B}}\right)\right)\right\Vert ^{2}\label{eq:wm}.
\end{align}
This search for permutations can be reduced to a classic linear assignment problem and be performed quickly. Equation~(\ref{eq:wm}) can be optimized using only weight without datasets. 

\noindent{\bf{STE:}} Straight Through Estimator (STE)\footnote{Note that while STE is typically an optimization method for non-differentiable problems, in this paper, it is used as the name of a model merging technique.} learns permutations to reduce the loss of parameters after merging as 
\begin{align}
&\min_{\mathbf{\tilde{w}_{B}}}\mathcal{L}_{AB}\left(\left(1-\lambda\right)\mathbf{w_{A}}+\lambda{\rm proj}\left(\mathbf{\tilde{w}_{B}}\right)\right),\nonumber\\&{\rm proj}\left(\mathbf{w}\right)=\argmin_{\pi\left(\mathbf{w_{B}}\right)}\left\Vert {\rm vec}\left(\mathbf{w}\right)-{\rm vec}\left(\pi\left(\mathbf{w_{B}}\right)\right)\right\Vert ^{2}.\label{eq:ste}
\end{align}
STE requires a dataset to estimate the loss. For merging models trained on a single dataset, Ainsworth~\citep{ainsworth2023git} reports that WM achieves similar accuracy to STE, despite not using the dataset. Therefore, WM was the main subject of investigation in their study~\citep{ainsworth2023git}.

\section{Model Merging between Different Datasets}\label{sec:modelnerge_ana}
To investigate model merging between different datasets, we first provide an intuitive understanding of model merging between different datasets and identify the causes of merging difficulties. Then, we show that datasets are required to effectively merge models between different datasets. Finally, we propose a model merging method using a surrogate dataset that decreases the requirement for the full dataset.

\begin{table}[tb]
    \centering
    \caption{\textbf{Effects of different datasets on WM.} L2 distance between $\mathbf{w_A}$ and $\mathbf{w_B}$ permuted using WM, along with the loss and accuracy of the merged model on the mixed dataset.}\label{tab:wm-differences}
    \small
\begin{tabular}{@{}ccccc@{}}
    \toprule
    \multicolumn{1}{c}{Degree} & \multicolumn{1}{c}{L2} & \multicolumn{1}{c}{Sharpness} & \multicolumn{1}{c}{Loss} & \multicolumn{1}{c}{Acc}  \\ \midrule
    $0^\circ$  & $3.84 \times 10^{-5}$ & $2.85 \times 10^{-4}$ & 0.008  & 98.49  \\
    $15^\circ$ & $3.96 \times 10^{-5}$ & $3.34 \times 10^{-4}$ & 0.062  & 97.53  \\    
    $30^\circ$ & $4.00 \times 10^{-5}$ & $3.42 \times 10^{-4}$ & 0.217  & 94.69  \\    
    $45^\circ$ & $4.05 \times 10^{-5}$ & $3.58 \times 10^{-4}$ & 0.390  & 89.68  \\    
    $60^\circ$ & $4.07 \times 10^{-5}$ & $3.77 \times 10^{-4}$ & 0.804  & 80.33  \\
    $75^\circ$ & $4.06 \times 10^{-5}$ & $3.72 \times 10^{-4}$ & 1.098  & 74.19  \\
    $90^\circ$ & $4.03 \times 10^{-5}$ & $3.65 \times 10^{-4}$ & 1.280  & 70.86  \\ \bottomrule
    \end{tabular}
\end{table}

\subsection{Causes of Difficulties with Model Merging}\label{sec:understand}
\begin{figure*}[tbh]
\centering
\begin{minipage}{0.67\linewidth}
    \begin{minipage}[t]{0.32\linewidth}
    \includegraphics[width=1\linewidth]{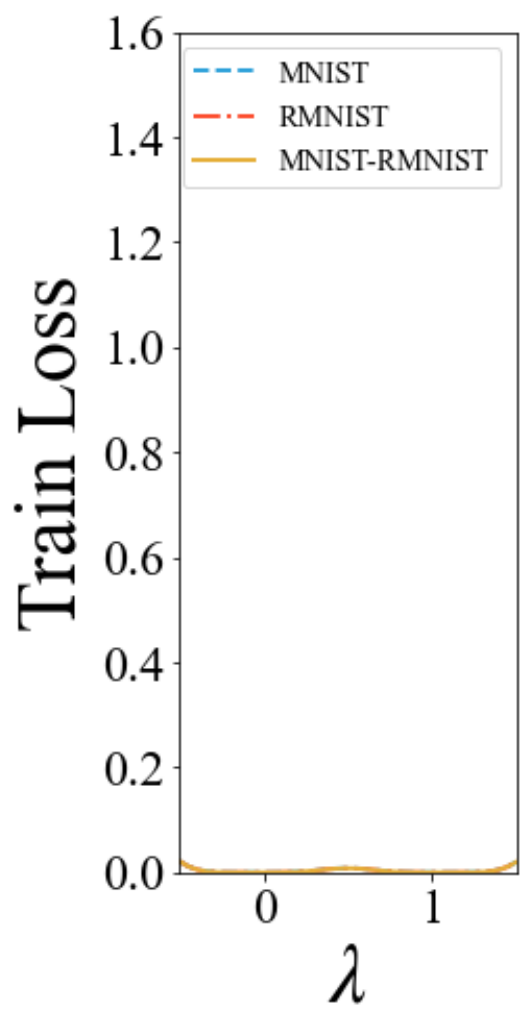}
    \subcaption{$0^\circ$~(single dataset)}\label{fig:losslandscape-rmnist-0}
    \end{minipage}  
    \begin{minipage}[t]{0.27\linewidth}\centering
    \includegraphics[width=1\linewidth]{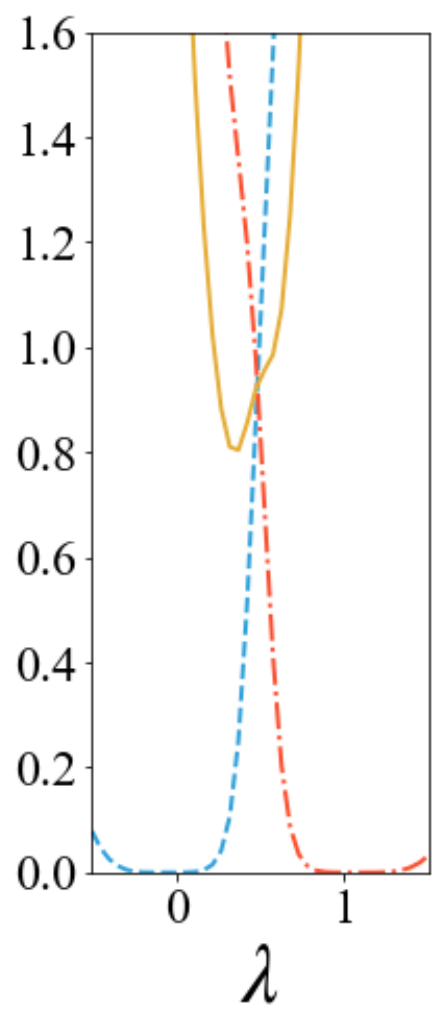}
    \subcaption{$45^\circ$}\label{fig:losslandscape-rmnist-45}
    \end{minipage}
    \begin{minipage}[t]{0.27\linewidth}
    \includegraphics[width=1\linewidth]{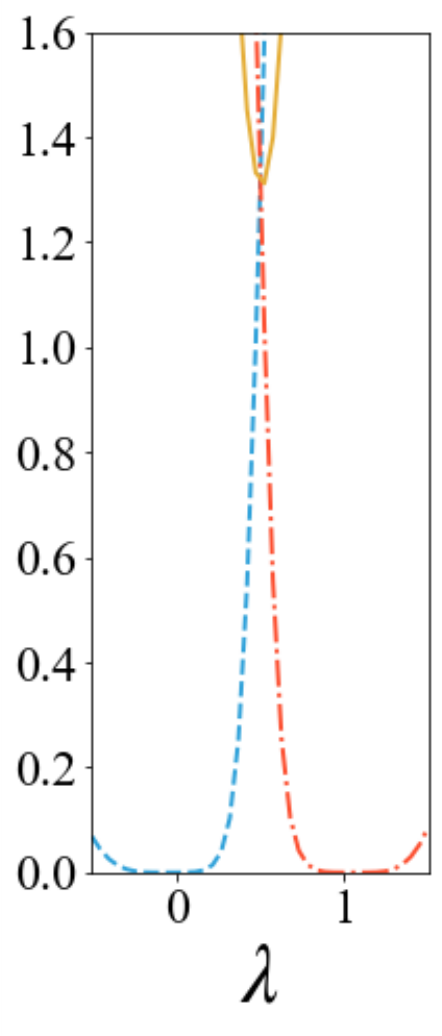}
    \subcaption{$90^\circ$}\label{fig:losslandscape-rmnist-90}
    \end{minipage}
    \end{minipage}
\begin{minipage}{0.25\linewidth}
        \begin{minipage}[t]{1\linewidth}
        \includegraphics[width=1\linewidth]{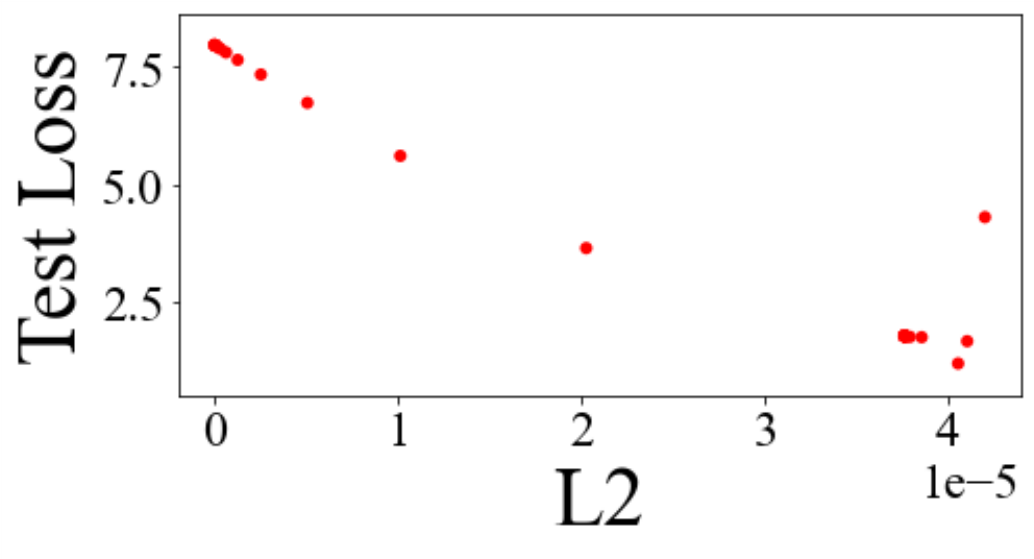}
        \end{minipage}  
        \begin{minipage}[t]{1\linewidth}\centering
        \includegraphics[width=1\linewidth]{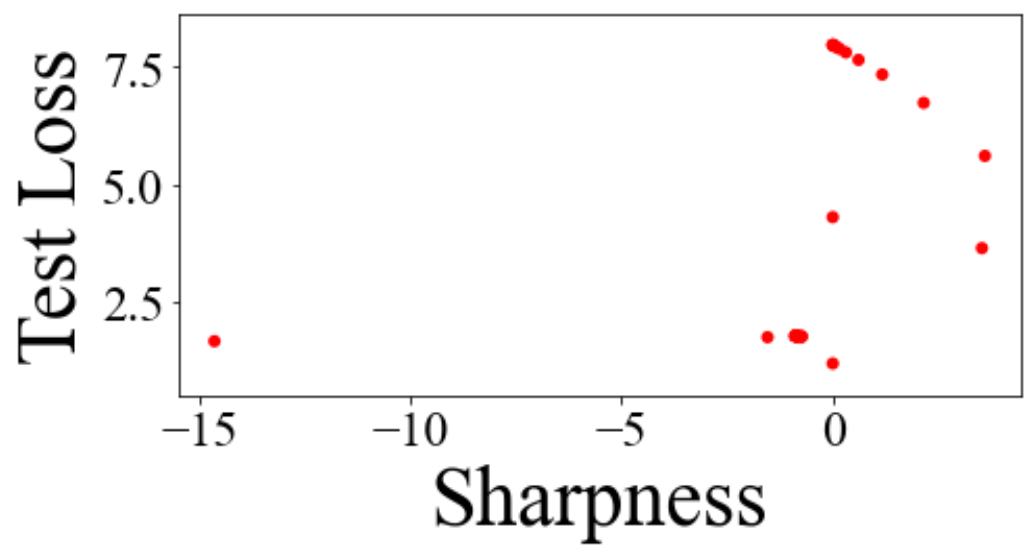}
        \end{minipage}
        \begin{minipage}[t]{1\linewidth}
        \includegraphics[width=1\linewidth]{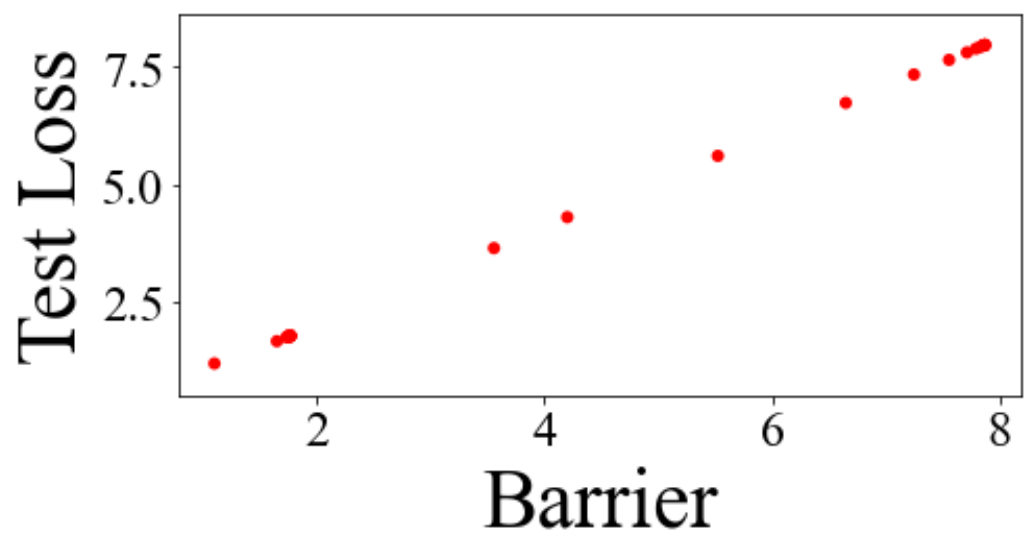}
        \end{minipage}
        \subcaption{Correlation}\label{fig:population}
\end{minipage}
\caption{\textbf{Impact of dataset differences on the loss landscape.} Figures~(\ref{fig:losslandscape-rmnist-0}, \ref{fig:losslandscape-rmnist-45}, and \ref{fig:losslandscape-rmnist-90}) visualize the loss on the mixed dataset in three different scenarios, each created by varying the degree of rotation, to explain the difficulty of model merging between different datasets. In each figure, we plotted training loss evaluated on three datasets (MNIST. RMNIST, MNIST-RMNIST) along the linear interpolation parameter $\lambda\in\left[0,1\right]$, with $\lambda=0$ representing the optimal weights for MNIST and $\lambda=1$ representing the optimal weights for RMNIST. Figure~(\ref{fig:population}) shows the relationship between L2 distance, sharpness, loss barrier, and test loss on a mixed dataset when various weight permutations are generated between MNIST and RMNIST-90. Reducing the L2 distance and sharpness does not necessarily decrease the test loss.
}\label{fig:losslandscape-rmnists}
\end{figure*}

To reveal why model merging between different datasets is difficult, we first present a theoretical intuition and then verify the intuition in an experiment.

\noindent{\bf{Theoretical intuition.}} The goal of model merging is to find $\mathbf{w_{AB}}$ that reduces loss $\mathcal{L}_{AB}$ from $\mathbf{w_A}$ and $\mathbf{w_B}$. To discuss the relationship between these weights, we first clarify the relationship between loss on different datasets. The loss on a mixed dataset, called loss barrier\footnote{This is a natural extension to the different datasets of the loss barrier on the single dataset~\citep{ainsworth2023git}. 
For more information, refer to the Sec.~\ref{sec:lossbarrier}.}, is as follows
\begin{align}
\mathcal{L}_{AB}\left(\mathbf{w}\right)=\frac{1}{2}\left(\mathcal{L}_{A}\left(\mathbf{w}\right)+\mathcal{L}_{B}\left(\mathbf{w}\right)\right)\label{eq:data_linear}
\end{align}
since the loss is the expected value $\mathcal{L}_{AB}=E_{P_{AB}\left(y,\mathbf{x}\right)}\left[-\log p\left(y|\mathbf{x},\mathbf{w}\right)\right]$ and can be separated for each dataset by using the linearity of the expectation. This equation shows that the loss landscape of $\mathcal{L}_{AB}$ is the sum of the loss landscapes of $\mathcal{L}_{A}$ and $\mathcal{L}_{B}$, as shown in Fig.~\ref{fig:losslandscape-rmnists}. We explain why WM has more difficulty merging models between different datasets than a single dataset. The permutation of WM brings $\mathbf{w_{B}}$ closer to $\mathbf{w_{A}}$. In scenarios of model merging between single datasets, $\mathcal{L}_{A}$ and $\mathcal{L}_{B}$ share optimal weights. Consequently, through WM, $\mathbf{w_{A}}$ converges towards $\mathbf{w_{B}}$, and setting these as $\mathbf{w_{AB}}$ results in a smaller loss. On the other hand, as the datasets become more divergent, the closest optimal weights have a large loss value on the datasets of each other, leading to an increased loss barrier.

\noindent{\bf{Experimental Design.}} To validate the above theoretical intuition, we show the loss landscape and the L2 distance between the closest optimal weights by continuously increasing the difference between Datasets A and B. We design Rotated-MNIST (RMNIST), which consists of MNIST images rotated by various angles, and conduct model merging experiments between MNIST and RMNIST. RMNIST uses rotations of $0^\circ$, $15^\circ$, $30^\circ$, $45^\circ$, $60^\circ$, $75^\circ$, and $90^\circ$. We train Model A on MNIST (Dataset A) and Model B on RMNIST (Dataset B), then use WM to permute$\mathcal{L}_{B}$, measuring L2 distance between closest optimal weights as shown in Table~\ref{tab:wm-differences} and Fig.~\ref{fig:losslandscape-rmnists}. 

\noindent{\bf{Experimental Verification.}}
Table~\ref{tab:wm-differences} shows that the L2 distance between the closest optimal weights of Models A and B diverges as the datasets become increasingly different, which leads to a decrease in the accuracy of the merged model. To directly verify this theoretical intuition, we visualize the loss landscape between the optimal weight of the MNIST and RMNIST in Fig.~\ref{fig:losslandscape-rmnists}. The blue, red, and yellow lines represent the loss landscapes on MNIST, RMNIST, and MNIST-RMNIST\footnote{We represent the pairs of datasets for model merging as [the dataset used to train Model A]-[the dataset used to train Model B].}, respectively. Figure~\ref{fig:losslandscape-rmnists} shows that the loss barrier (yellow line) increases as the angle of MNIST-RMNIST becomes more different.


\subsection{Conditions for Successful Merging}\label{sec:merge-condition} 
Figures~\ref{fig:losslandscape-rmnist-45} and ~\ref{fig:losslandscape-rmnist-90} show WM does not make these optimal weights close enough to ignore the increase in loss. This suggests that a simple WM approach, focusing only on minimizing the L2 distance, is insufficient for effectively merging models between different datasets. To address this challenge, we investigate three types of information related to model merging to identify the information required for successfully merging models: (i) L2 distance between weights (Eq.~\ref{eq:wm}), (ii) sharpness, and (iii) loss barrier (Eq.~\ref{eq:data_linear}), where the sharpness between $\mathbf{w_{A}}$ and $\mathbf{w_{B}}$ is defined as
\begin{align}
\frac{1}{2}\left[\left(\!\mathbf{w_{B}}\!-\!\mathbf{w_{A}}\!\right)^{\top}\frac{\partial\mathcal{L}_{A}\left(\!\mathbf{w_{A}}\!\right)}{\partial\mathbf{w}}\!+\!\left(\!\mathbf{w_{A}}\!-\!\mathbf{w_{B}}\!\right)^{\top}\frac{\partial\mathcal{L}_{B}\left(\!\mathbf{w_{B}}\!\right)}{\partial\mathbf{w}}\right].
\end{align}
As shown in Fig.~\ref{fig:losslandscape-rmnist-90}, the loss barrier is determined by both the L2 distance and the degree of increase in loss. Therefore, to represent the information about the increase in loss, we introduce the concept of sharpness.
Fig.~\ref{fig:population} shows the relationship between L2 distance, sharpness, loss barrier, and test loss on a mixed dataset when various weight permutations are generated between MNIST and RMNIST-90 (for the detailed experimental setup in Sec.~\ref{sec:population_setting}). The use of various permutations helps to avoid dependency on the search methodology. Figure~\ref{fig:population} indicates that the test loss does not become smaller as the L2 distance becomes smaller. 
Figure~\ref{fig:population} also shows that the sharpness is poorly correlated with test loss of the merged model. This is because loss barriers that deviate from the optimal weights become difficult to estimate when using local gradients on the optimal weights\footnote{In fact, we tried a method using regularization to reduce sharpness for WM, but it did little to improve accuracy 
(see Sec.~\ref{sec:flat_weight_matching}).}. 
On the other hand, the loss barrier, which is training loss on mixed datasets, correlates well with test loss\footnote{It is checked for a generalization gap.}. 
These experimental results indicate that directly reducing the loss barrier is most likely to lead to successful model merging when different datasets are used. Since directly reducing the loss barrier requires the use of STE with datasets to be applied, datasets are currently required for efficient model merging. We will show that STE works effectively between different datasets in Sec.~\ref{sec:experiments}.

\subsection{Model Merging with Surrogate Datasets}
\vspace{-0.5em}
Model merging between different datasets requires a smaller loss barrier by using the datasets. However, this is a major constraint on sharing all data in terms of heterogeneous data utilization, data privacy, and data storage cost. To relax this constraint, we propose a method that uses a surrogate dataset instead of a real dataset. We employ two methods to create the surrogate dataset: coreset selection and dataset condensation.

Coreset selection aims at reducing the size of a dataset by selecting a smaller, representative group of data points, with the goal of approximating training on the full dataset. Various coreset selection methods have been proposed, but when an extremely small coreset is selected, the performance is almost identical to that of random sampling~\citep{guo2022deepcore}. We experimented with five methods, including random sampling, for generating surrogate datasets and merging models with these surrogate datasets and found that the accuracy was the same as that of random sampling 
(see Sec.~\ref{sec:modelmerge-coreset}). 
Therefore, we will primarily discuss random sampling in the rest of this paper. 

Dataset condensation is a method for condensing a large dataset into a small set of informative synthetic samples for training DNNs from scratch. Various methods have been proposed for dataset condensation, and we use a major method called Gradient Matching~\citep{zhao2021dataset} in this paper. Gradient Matching creates a condensed dataset so that the gradients of the weights trained on the original dataset match the gradients of the condensed dataset. Dataset condensation is more complex than coreset selection using random sampling. However, with more sophisticated methods~\citep{guo2023towards,nguyen2021dataset} being proposed recently, there is potential for significant improvements in accuracy in the future.

To summarize the overall procedure for model merging without the full dataset, we first prepare trained models for $P_A$ and $P_B$ and then perform coreset selection and dataset condensation on $P_A$ and $P_B$ to create a surrogate dataset. Next, we mix surrogate Datasets A and B to create surrogate Dataset AB. Finally, we merge models with STE using surrogate Dataset AB instead of the full dataset $P_{AB}$.


\section{Experiments}\label{sec:experiments}
\noindent{\bf{Outline.}}
This section experimentally confirms that STE is effective for model merging between different datasets and that a reasonable permutation can be achieved even using STE with surrogate datasets. The basic experimental procedure is to first train independently on $P_{A}$ and $P_{B}$ to obtain $\mathbf{w_{A}}$ and $\mathbf{w_{B}}$. Then the weights $\mathbf{w_{B}}$ are permuted. Finally, model parameters are merged as $\mathbf{w_{AB}}=\left(1-\lambda\right)\mathbf{w_{A}}+\lambda\pi\left(\mathbf{w_{B}}\right)$ and sliding $\lambda$ to evaluate the accuracy on $P_{AB}$.

\noindent{\bf{Datasets.}} 
First, we conduct model merging between different datasets with shared labels, such as MNIST-RMNIST and USPS-MNIST. Both MNIST~\citep{lecun1998gradient} and USPS~\citep{hull1994database} are grey-scaled character datasets. Originally, Rotated-MNIST~(RMNIST) is a dataset with MNIST rotated by an arbitrary angle, but in this section, we set the rotation angle of RMNIST to $90^\circ$. Next, we experiment with the case of no label sharing as SPLIT-CIFAR10 and MNIST-FMNIST. SPLIT-CIFAR10 divides CIFAR10~\citep{krizhevsky2009learning} into two datasets, one with labels 0-4 and the other with labels 5-9.
MNIST-FMNIST merges models between two completely different datasets: MNIST, which includes digit images, and Fashion-MNIST~(FMNIST)~\citep{xiao2017online}, which includes fashion images\footnote{Both MNIST and FMNIST have 10 classes, and to merge with all weights, including up to the last layer, class 0 in MNIST and class T-shirt/top in FMNIST mean Label 0.}.

\noindent{\bf{Setup.}}
We use the same setup for the preprocessing and network architecture as the existing method~\citep{ainsworth2023git}. MNIST, USPS, RMNIST, and FMNIST use a fully connected network, and CIFAR10 uses ResNet20$\times16$, which is 16 times wider than ResNet20. This is because existing research has shown that model merging on a single dataset in CIFAR10 cannot be effective unless the width is sufficiently large. Since Batch Normalization is also known to destroy the LMC, we use REPAIR (REnormalizing Permuted Activations for Interpolation Repair)~\citep{jordan2022repair} to avoid this. We provide the results of model merging on a single dataset as a preliminary experiment in 
Sec.~\ref{sec:same-data}. 
To make surrogate datasets, we applied both coreset selection~(Coreset) and dataset condensation~(Data Cond) techniques. For Coreset, we show that random sampling is the best choice for the coreset selection in 
Sec.~\ref{sec:modelmerge-coreset}. 
As for Data Cond, we use Gradient Matching~\citep{zhao2021dataset} for USPS, MNIST, RMNIST, and FMNIST, and public datasets\footnote{\href{https://github.com/google-research/google-research/tree/master/kip}{github.com/google-research/google-research/tree/master/kip}} using sophisticated methods~\citep{nguyen2021dataset} for SPLIT-CIFAR10. We use 10 images per class for USPS, MNIST, RMNIST, and FMNIST, and 50 images per class for SPLIT-CIFAR10. For full experimental details in Sec.~\ref{sec:experiments_details}.

\noindent{\bf{Baselines.}} 
We compare the five methods as a baseline for model merging. \textbf{Na\"ive}: Simple average method as $\mathbf{w_{AB}}=\left(1-\lambda\right)\mathbf{w_{A}}+\lambda\mathbf{w_{B}}$. \textbf{WM}: The weights are permuted to close the L2 distance between the weights of the two models and then averaged. \textbf{ZipIt!}: ZipIt!~\citep{stoica2023zipit} merges models using correlated neurons, which allows for the combination of neurons both between different networks and within the same network. We use the full dataset to estimate correlated neurons. \textbf{OT}: Model Fusion via OT~\citep{singh2020model} is sophisticated model merging, with a higher degree of freedom than weights permutation. \textbf{OT~(Tuned)}: It uses $\lambda$ with the highest test accuracy from $\lambda\in\left[0,1\right]$. Since there is a large divergence in performance between $\lambda=0.5$ and Tuned for OT, we include both, while the other methods include results for $\lambda=0.5$. 
We also compare with another non-merging baseline due to its similar problem setting to model merging. \textbf{Ensemble}: It averages the predictions of the model. Note that ensemble is compared as a reference due to the doubled inference cost.

\subsection{Model Merging between Different Datasets.}
\vspace{-0.5em}
\begin{figure*}
\centering
\begin{minipage}[t]{0.325\linewidth}
\includegraphics[width=1\linewidth]{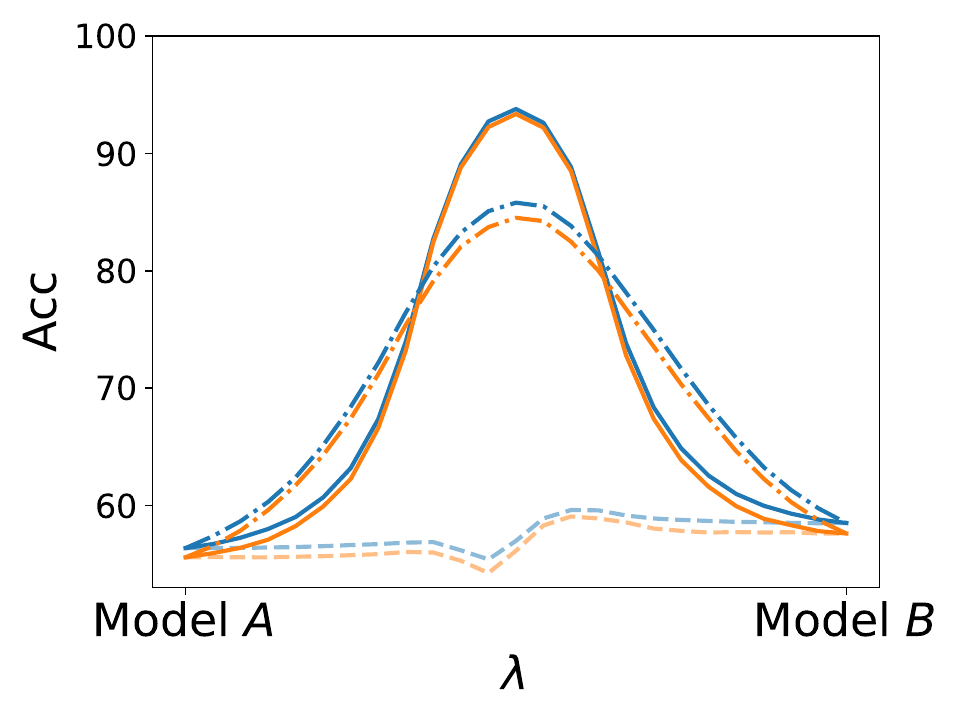}
\subcaption{STE~(Full Data)}\label{fig:rotated-mnist-full}
\end{minipage}
\begin{minipage}[t]{0.325\linewidth}
\includegraphics[width=1\linewidth]{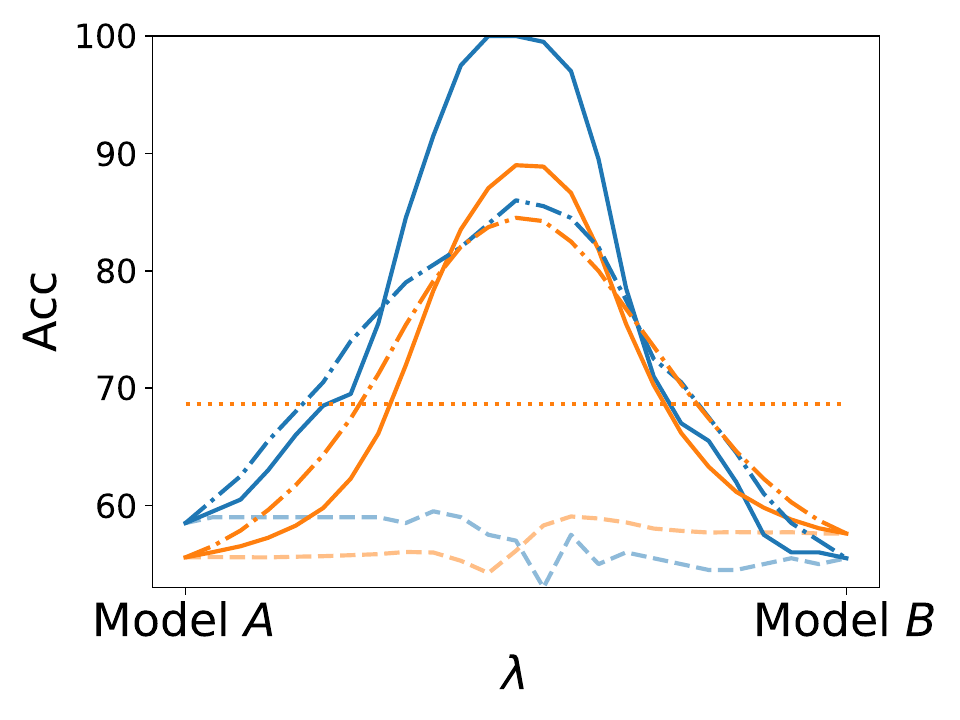}
\subcaption{STE~(Coreset)}\label{fig:rotated-mnist-coreset-10}
\end{minipage}
\begin{minipage}[t]{0.325\linewidth}\centering
\includegraphics[width=1\linewidth]{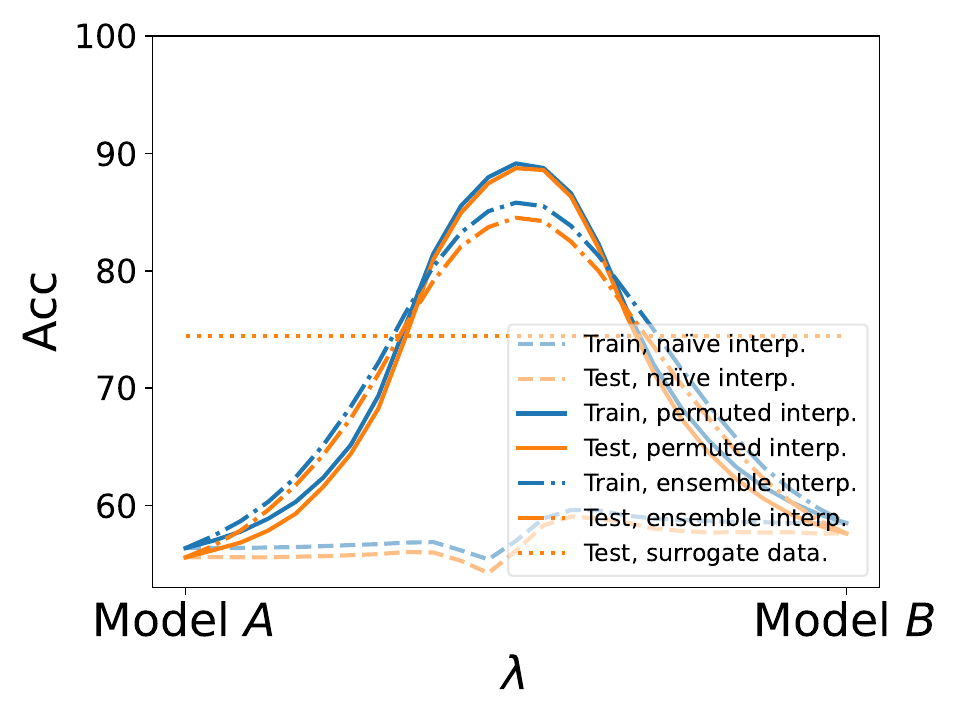}
\subcaption{STE~(Data Cond)}\label{fig:rotated-mnist-cond-10}
\end{minipage}
\caption{\textbf{Model merging between different datasets is possible after permuting even with significantly small surrogate datasets.} Accuracy interpolating between models trained on MNIST and RMNIST. The vertical axis represents accuracy on the mixed dataset MNIST-RMNIST. The solid orange line represents the test accuracy of the merged model using STE. Figure~\ref{fig:rotated-mnist-full} shows model merging with full data. Figures~\ref{fig:rotated-mnist-coreset-10} and \ref{fig:rotated-mnist-cond-10} show merging with only 10 samples per class, using coreset selection and dataset condensation. The horizontal dashed lines represent Coreset AB and Data Cond AB methods.}
\label{fig:rotated-mnist}
\end{figure*}

First, for an intuitive understanding of our experiments, we present the model merging between MNIST and RMNIST. MNIST is Dataset A and RMNIST is Dataset B, and the models trained on MNIST and RMNIST are denoted as Models A and B, respectively. Figure~\ref{fig:rotated-mnist} shows the accuracy of the merged model on $P_{AB}$, which is merged by STE and STE with dataset condensation. Model A~($\lambda=0$) has an accuracy of around $55\%$ on the MNIST and RMNIST mixed dataset, since it is almost $100\%$ accurate on MNIST and $10\%$ on RMNIST. Model B~($\lambda=1$) is similar. As shown in Fig.~\ref{fig:rotated-mnist-full}, the permuted~(STE) achieves $93\%$ of test accuracy compared to $56\%$ for Na\"ive. This is surprising as there is an effective weight for model merging in the permutation of neurons.

\begin{table*}[tb]
\caption{\textbf{Test accuracies~(\%) on $D_{AB}$ for various model merging between different dataset.} In model merging between different datasets, the results demonstrate that both STE and STE with surrogate datasets achieve higher accuracy than other methods of merging models. We omitted the error bars from five seeds as they are sufficiently small, being less than 1\%.}\label{tab:summary}
\small
\centering  
\begin{tabular}{@{}lcccc@{}}
\toprule
                   & MNIST-RMNIST & USPS-MNIST & SPLIT-CIFAR10 & MNIST-FMNIST \\ \midrule
Model A            & 55.58        & 76.04      & 48.40         & 53.21        \\
Model B            & 57.61        & 62.99      & 48.90         & 51.81        \\
Model AB           & 98.19        & 96.92      & 95.63         & 94.03       \\ \midrule
Ensemble           & 84.51        & 93.46      & 79.79         & 75.93        \\
Data Cond AB       & 74.47        & 82.55      & 39.54         & 73.28        \\
Coreset AB         & 67.77        & 76.69      & 38.50         & 66.03        \\ \midrule
Naïve              & 56.14        & 68.85      & 10.00         & 45.75        \\
WM                 & 70.92        & 84.21      & 58.91         & 55.61        \\
OT                 & 37.06        & 79.20      & 24.01         & 24.76        \\
OT~(Tuned)         & 58.57        & 82.39      & 46.62         & 48.10        \\
ZipIt!             & 71.35        & 79.96      & 72.05         & 58.95        \\
STE~(Full)         & 93.52        & 95.50      & 89.80         & 83.76        \\
STE~(Coreset)      & 88.78        & 93.30      & 82.60         & 77.77        \\ 
STE~(Data Cond)    & 88.78        & 92.50      & 71.95         & 80.94        \\\bottomrule
\end{tabular}
\end{table*}

Next, similar experiments are conducted across the various datasets. The results are shown in Tab.~\ref{tab:summary}. In the table, \textbf{Model AB} is the oracle model trained with the mixed dataset, while Models A and B are the pre-merging models trained on Datasets A and B, respectively. 
 STE~(Full) outperforms all other baseline methods and ensemble. STE~(Full) achieves high accuracy, especially for merging, which is difficult even with OT and ZipIt! such as MNIST-FMNIST. For model merging on a single dataset, WM and STE~(Full) are reported to perform similarly~\citep{ainsworth2023git}. However, in model merging between different datasets, we find that STE~(Full) significantly outperforms WM. This implies that not only the weights but also the information about the datasets is important when merging between different datasets.

\subsection{Model Merging with Surrogate Datasets}
\vspace{-0.5em}
\begin{table}[tbh]
\small
\caption{\textbf{Impact of surrogate datasets sample size.} It compares test accuracy for STE~(Coreset) and Corset AB on SPLIT-CIFAR10 with varying numbers of samples.
Columns represent the number of samples per class.}\label{tab:number-samples}
\centering
\begin{tabular}{@{}llllll@{}}
\toprule
              & 1     & 10    & 50    & 100   & 500  \\ \midrule
STE (Coreset) & 61.15 & 75.13 & 82.77 & 84.85 & 88.62\\
Coreset AB     & 19.32 & 25.53 & 41.61 & 49.31 & 76.47\\ \bottomrule
\end{tabular}
\end{table}

Through our experiments with surrogate datasets, we demonstrate that model merging is effective even with a reduced number of data points. We compare our method to STE~(Coreset) and STE~(Data Cond). \textbf{STE~(Coreset)} uses a surrogate datasets made with coreset selection, and \textbf{STE~(Data Cond)} uses one made with dataset condensation. We find that the proposed method significantly outperforms the na\"ive method, even with as few as 10 data points per class (approximating $0.2{\rm \%}$ of the training dataset). This suggests a notable efficiency in model merging when dealing with limited data. In comparison to other baselines, STE~(Coreset) and STE~(Data Cond) achieve higher accuracy than the other baseline methods except for Ensemble, which has a high inference cost. It is particularly surprising that coreset selection, even with a small number of data points chosen through simple random sampling, is sufficient. 

To assess the effectiveness of model merging in scenarios with limited data points, we compare our method with the \textbf{Coreset AB} and \textbf{Data Cond AB} methods, which are each trained from scratch on surrogate Dataset AB, created through coreset selection and dataset condensation, respectively. This comparison is crucial because if effective performance is achieved with training from scratch on the smaller surrogate dataset of $P_{AB}$, it could suggest that the additional benefits of model merging are limited. Despite this possibility, our method demonstrates superior accuracy over both the Coreset AB and Data Cond AB methods. This advantage comes from its ability to align the weights and maintain optimized loss values for each dataset, resulting in lower loss values on the mixed dataset. Table~\ref{tab:number-samples} shows the impact of the number of surrogate datasets samples on model merging. In all cases, STE~(Coreset) outperforms Coreset AB. These findings are further supported by the results in Figs.\ref{fig:rotated-mnist-coreset-10} and \ref{fig:rotated-mnist-cond-10} and are comprehensively summarized in Table\ref{tab:summary}.

\subsection{Computational Cost of Mode Merging}
\vspace{-0.5em}
\begin{figure}[tbh]
\centering
\includegraphics[width=0.5\linewidth]{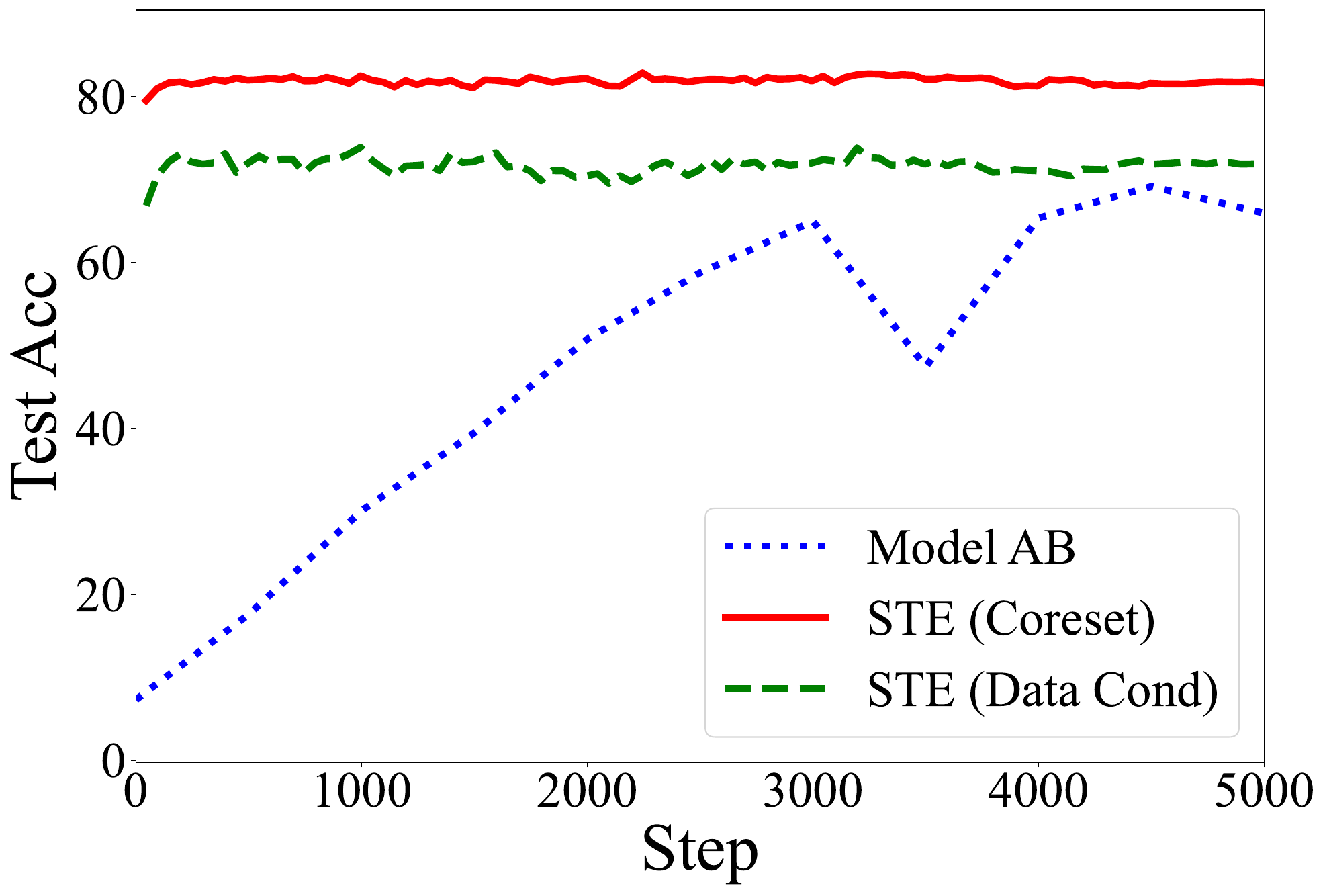}
\caption{\textbf{STE with surrogate datasets is much faster than full training.} This Table shows a comparison of computational costs: full training vs. model merging. The vertical axis represents the test accuracy on Dataset AB. The horizontal axis shows the number of steps required both for training from scratch on Dataset AB and for merging models trained on Datasets A and B. To ensure a fair comparison, training and merging are performed with the same batch size.}
\label{fig:computationalcost}
\end{figure}
In this section, we evaluate the computational cost of the model merging. Our comparison includes three models: Model AB, STE~(Data Cond), and STE~(Coreset), chosen for their demonstrated high accuracy in Table~\ref{tab:summary}. 
Figure~\ref{fig:computationalcost} shows test accuracy and the number of steps for training and merging on the SPLIT-CIFAR10. Model merging using surrogate datasets achieves higher accuracy with fewer steps than Model AB, which is trained on the mixed dataset.

\section{Related Work}
\vspace{-0.5em}
First, we introduce existing methods that have a similar purpose to model merging and organize the differences with model merging. Then, we introduce LMC, which is closely related to model merging. Finally, the differences with existing model merging research are summarized.

\noindent{\bf{Continuous learning.}} 
Continuous learning is a method for learning new knowledge while retaining previously acquired knowledge~\citep{silver2013lifelong}. A major strategy in continuous learning is to separate the weights used for learning between Datasets A and B~\citep{kirkpatrick2017overcoming,mallya2018packnet}. 

\noindent{\bf{Federated learning.}}
Federated learning is a learning method without directly sharing the datasets in cases where each training dataset is distributed~\citep{mcmahan2017communication}. Federated learning is promising in terms of data privacy, security, and the use of heterogeneous data, since datasets are not shared. Instead of sharing data points, gradients of weights for losses are shared. Like model merging with condensed datasets, federated learning aims to create a model that adapts to the entire dataset without the need to share the raw data.

Continuous learning and federated learning often involve a dependency where the training of Model B relies on Model A, limiting adaptability to pre-trained models. Conversely, model merging allows flexible combinations with other models like Model C without retraining. This is because model merging is conducted independently after the training of Models A and B.

\noindent{\bf{Ensemble.}}
Ensemble methods~\citep{breiman1996bagging,wolpert1992stacked,schapire1999brief} improve the accuracy by taking the average of the predictions of several different models. 
As was shown in Sec.~\ref{sec:experiments}, the ensemble method works surprisingly well for merging between different datasets. However, ensemble methods have disadvantages in terms of computational cost and memory efficiency since they use the predictions of all models in inference. 

\noindent{\bf{LMC.}} The concept of mode connection, where the optimal weights of the scholastic gradient descent (SGD) have connected paths between each other, was introduced by Timur~\citep{garipov2018loss}. Furthermore, Jonathan~\citep{frankle2020linear} proposed the LMC and found a relationship between the LMC and the Lottery Hypothesis~\citep{frankle2018the}. It has been reported that LMC is valid if the neurons of two networks trained with different random seeds are appropriately permuted~\citep{entezari2021role,ainsworth2023git}. Furthermore, LMC has been reported to be valid between fine-tuning models trained with different random seeds from a single pre-training model~\citep{NEURIPS2020_0607f4c7}. Thus, LMC has attracted attention as a tool for understanding SGD, the Lottery Hypothesis, and fine-tuning.

\noindent{\bf{Merging Methods.}} 
Model merging can be categorized into two scenarios: FT-Merge and Pretrain-Merge.
{\bf{FT-Merge}} is a specific scenario where models fine-tuned from a same pre-trained model are merged. In the case of FT-Merge, merging can be done through weight averaging or similar methods, including Gradient Matching~\citep{daheim2024model}, Elastic Weight Consolidation~\citep{leontev2020non} and Fisher-Weighted Averaging~\citep{matena2022merging}, as implemented in mergekit~\citep{goddard2024arcee}\footnote{Excluding Elastic Weight Consolidation.}. This is possible for FT-Merge because, being fine-tuned from the same initial weights with a small learning rate, the two fine-tuned models exist in the same loss basin. Therefore, there is no loss barrier between the two models, allowing the weight averaging strategy to work well.
{\bf{Pretrain-Merge}} (our problem setting) is a method that merges models trained from arbitrary initial values, not constrained by fine-tuning. Pretrain-Merge aims to merge models from different initial values, and it does not have the constraints present in FT-Merge, i.e., sharing the same pre-trained model. Pretrain-Merge methods using optimal transport (OT)~\citep{singh2020model} and using correlation of neurons~\citep{stoica2023zipit} have been proposed. We compare these methods as baselines in our experiment.


\section{Limitations}
Our results have several limitations. (1) The analysis in Sec.~\ref{sec:modelnerge_ana} assumes the existence of optimal weights $\mathbf{w_{A}}$ and $\mathbf{w_{B}}$, which are close enough to be considered convex functions, as shown in Fig.~\ref{fig:losslandscape-rmnists}. (2) We showed that STE can reach effective permutations experimentally, but there is a large number of permutation patterns, thus there is no theoretical guarantee, and theoretical analysis is a remaining task. (3) Our experiments are limited to image classification between the same model architectures. Adaptation of model merging to natural language datasets and generative models is an intriguing application (e.g. Checkpoint Merger of Stable Diffusion web UI~\citep{automatic1111}).
\section{Conclusion and Future Work}
This paper investigated model merging between different datasets. We found that there is an effective weight for model merging in the permutation of neurons. This finding is a motivation for future research to improve the efficiency of permutation search for model merging between different datasets. Furthermore, we proposed model merging by using a surrogate dataset, without sharing all data points. Future work needs to examine whether a surrogate dataset protects privacy. In other research directions, the model merging between many models is of interest. In DNN training, huge computational costs are being spent to develop huge models to improve accuracy, but this may be a limitation from the viewpoint of system energy consumption. Efficient merging of multiple models could pave the way for new learning methods as~\citep{li2022branch} to replace current learning methods where a single data point affects all parameter updates.

\bibliography{modelmerge}
\bibliographystyle{acml24}

\appendix
\section{Experimental Details}\label{sec:experiments_details}
\subsection{Dataset Condensation}
For MNIST, RMNIST, FMNIST, and USPS we condensed the dataset using Gradient Matching~\citep{zhao2021dataset}. The setup for creating the condensed dataset was the same architecture and the same data preprocessing as used for training. Condensation was performed in two cases: 1 per label and 10 per label. 
In CIFAR10, we used a more sophisticated dataset condensation~\citep{nguyen2021dataset}. Since compressed datasets were publicly available\footnote{\url{https://storage.cloud.google.com/kip-datasets/kip/cifar10/ConvNet_ssize500_nozca_nol_noaug_ckpt11500.npz}}, we used them.

\subsection{Setup in Fig.~\ref{fig:population}}
\label{sec:population_setting}
Can we find the effective permutation for model merging without data? To answer this question, we generated a variety of permutations and plotted the correlations between test loss on the mixed dataset and neuronal permutation criteria in Fig.~\ref{fig:population}. The datasets used were MNIST and RMNIST-90. To produce permutation variations, we used $25$ patterns of FWM $\beta$ from $0$ to $1$ and max iteration of WM search from $\left[2,5,300\right]$ and random number seed $\left[0,1,2,3,4\right]$. Of the 375 permutations generated, 30 are unique.

\subsection{Setup of Tab.~\ref{tab:summary}}
To test the stability of the model merging method, we loaded the checkpoints of the trained Models A and B and experimented with different random number seeds. We merge the models five times to show the average test accuracy. The error bars are small enough to be omitted.

\subsection{Hardware and software}
Our SPLIT-CIFAR10 experiments are done on a workstation with Intel Xeon(R) Gold 6128 CPU 8 cores, 128GB RAM, and 4 NVIDIA Tesla V100 GPUs. Other experiments are done on M1 max Mac Book Pro. The deep learning models are implemented in PyTorch~\citep{NEURIPS2019_9015}.
\subsection{Rotated MNIST}
When training without data augmentation, models trained on MNIST will have lower accuracy on RMNIST. Since our implementation of OT~\citep{singh2020model} could not handle the bias term in the fully connected layer, the RMNIST experiment used the same checkpoints as the proposed method with the bias term turned off after loading. We found that turning off the bias term did not change the accuracy of the model much: test acc was $98.45$ with the MNIST-0 bias term, $98.11$ without it, $98.33$ with the MNIST-90 bias term, and $98.19$ without it.

\subsection{SPLIT-CIFAR10}
Since our implementation of OT could not handle the bias and batch normalization terms in the fully connected layer, we evaluated OT by training on the same network as STE with the Bias and Batch Normalization turned off. This is because when we experimented with the same checkpoint as the proposed method with the bias term and batch normalization turned off, the accuracy of the individual models before merging degraded significantly. Turning off the batch norm and bias affects the accuracy of Model A on $D_{AB}$ to $48.40\rightarrow44.09$ and Model B to $48.90\rightarrow46.62$. In OT, act-num-samples were set to $10$ due to memory.

\section{Theoretical Analysis}
\label{sec:direction}
\subsection{Do Optimal Weights of Dataset AB Lie on a Line Segment between Optimal Weights of Datasets A and B?}
\begin{figure*}[b]
\centering
\includegraphics[width=0.4\linewidth]{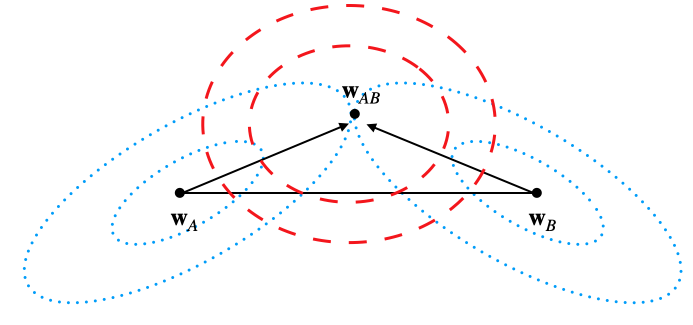}
\caption{2D Contours Plot. The relationship between the optimal weights on Datasets A and B before merging and the optimal weights on the optimal datasets after merging. The blue lines represent the contours plot of the loss on Datasets A and B, and the red lines represent the contours plot of the loss on Dataset AB after merging.}\label{fig:hessian}
\end{figure*}

In this subsection, we discuss whether the optimal weights in a mixed dataset are on the line between $\mathbf{w_{A}}$ and $\mathbf{w_{B}}$. Let us assume that weight permutation makes $\mathbf{w_{A}}$ and $\mathbf{w_{B}}$ close enough $\mathbf{w}\in\sigma\left(\mathbf{w_{A}},\mathbf{w_{B}}\right)$ and that their respective losses are in a region where they can be regarded as convex functions. This assumption holds with permutation by WM and STE. This is confirmed in Sec.~\ref{sec:losslandscape}. In Eq.~(\ref{eq:data_linear}), $\mathcal{L}_{AB}$ is a convex function because it is a convex combination of convex functions $\mathcal{L}_{A}$ and $\mathcal{L}_{B}$, and it is expressed as a quadratic equation around the optimal weight $\mathbf{w_{AB}}$. If we expand both sides of Eq.~(\ref{eq:data_linear}), we obtain
\begin{align}
\mathcal{L}_{AB}\left(\mathbf{w_{AB}}\right)+\left(\mathbf{w}-\mathbf{w_{AB}}\right)F_{AB}\left(\mathbf{w}-\mathbf{w_{AB}}\right)&=\frac{1}{2}\left(\mathcal{L}_{A}\left(\mathbf{w_{A}}\right)+\left(\mathbf{w}-\mathbf{w_{A}}\right)F_{A}\left(\mathbf{w}-\mathbf{w_{A}}\right)\right.\nonumber\\&\left.+\mathcal{L}_{B}\left(\mathbf{w_{B}}\right)+\left(\mathbf{w}-\mathbf{w_{B}}\right)F_{B}\left(\mathbf{w}-\mathbf{w_{B}}\right)\right),
\end{align}
where $F$ is Fisher information matrix and the first derivative is zero due to the optimal weights. A coefficient comparison of both sides shows that
\begin{align}
\mathbf{w_{AB}}=\frac{1}{2}\left(F_{AB}\right)^{-1}\left(F_{A}\mathbf{w_{A}}+F_{B}\mathbf{w_{B}}\right),\qquad F_{AB}=F_{A}+F_{B}\label{eq:direction}.
\end{align}
This means that the optimal direction of the weights of $\mathcal{L}_{AB}$ is determined from the Hesse matrices for the respective losses of $\mathcal{L}_{A}$ and $\mathcal{L}_{B}$. Figure~\ref{fig:hessian} shows that the relationship between the loss landscapes of Datasets A, B, and AB. If $F_{A}$ and $F_{B}$ are diagonal, i.e., the loss contours are isotropic, then $\mathbf{w_{AB}}$ lies on the linear connection of $\mathbf{w_{A}}$ and $\mathbf{w_{B}}$. In practice, the optimal weights $\mathbf{w_{AB}}$ lie out of linear connection due to the distortion of the loss planes. The optimal weights are found to be in the direction of the eigenvectors corresponding to the smaller eigenvalues of the Fisher information matrix as Fig.~\ref{fig:hessian}. 

\subsection{Derivation of Eq.~(\ref{eq:direction})}
\label{sec:drive_direction}
Let us assume that $\mathbf{w_{AB}}$, $\mathbf{w_{A}}$ and $\mathbf{w_{B}}$ are close enough $\mathbf{w}\in\sigma\left(\mathbf{w_{AB},w_{A}},\mathbf{w_{B}}\right)$ and that their respective losses are in a region where they can be regarded as convex functions. We expand both sides of Eq.~(\ref{eq:data_linear}) as
\begin{align}
\mathcal{L}_{AB}\left(\mathbf{w_{AB}}\right)+\left(\mathbf{w}-\mathbf{w_{AB}}\right)F_{AB}\left(\mathbf{w}-\mathbf{w_{AB}}\right)&=\frac{1}{2}\left(\mathcal{L}_{A}\left(\mathbf{w_{A}}\right)+\left(\mathbf{w}-\mathbf{w_{A}}\right)F_{A}\left(\mathbf{w}-\mathbf{w_{A}}\right)\right.\nonumber\\&\left.+\mathcal{L}_{B}\left(\mathbf{w_{B}}\right)+\left(\mathbf{w}-\mathbf{w_{B}}\right)F_{B}\left(\mathbf{w}-\mathbf{w_{B}}\right)\right).
\end{align}
A comparison of the coefficients on both sides shows that
\begin{align}
\mathbf{w}F_{AB}\mathbf{w}=\mathbf{w}\frac{1}{2}\left(F_{A}+F_{B}\right)\mathbf{w}
\end{align}
\begin{align}
-\mathbf{w}F_{AB}\mathbf{w_{AB}}-\mathbf{w_{AB}}F_{AB}\mathbf{w}&=\frac{1}{2}\left(-\mathbf{w}F_{A}\mathbf{w_{A}}-\mathbf{w_{A}}F_{A}\mathbf{w}-\mathbf{w}F_{B}\mathbf{w_{B}}-\mathbf{w_{B}}F_{B}\mathbf{w}\right)\nonumber\\\mathbf{w}F_{AB}\mathbf{w_{AB}}&=\mathbf{w}\frac{1}{2}\left(F_{A}\mathbf{w_{A}}+\mathbf{w}F_{B}\mathbf{w_{B}}\right).
\end{align}
We used the Fisher information matrix as a symmetric matrix. We can obtain $\mathbf{w_{AB}}=\frac{1}{2}\left(F_{AB}\right)^{-1}\left(F_{A}\mathbf{w_{A}}+F_{B}\mathbf{w_{B}}\right)$ and $F_{AB}=\frac{1}{2}\left(F_{A}+F_{B}\right)$.

\subsection{Losses on Mixed Datasets}
In this subsection, we show that the average of the $\mathbf{w_{A}}$ and the permuted $\mathbf{w_{B}}$ keeps losses on $\mathcal{L}_{AB}$ low as
\begin{align}
\mathcal{L}_{AB}\left(\mathbf{w}_{AB}\right)=\mathcal{L}_{AB}\left(\frac{1}{2}\left(\mathbf{w_{A}}+\mathbf{w_{B}}\right)\right)\label{eq:loss_val}.
\end{align}
To understand model merging intuitively, the $\mathbf{w_{AB}}$ and $\mathbf{w_{A}}$ and $\mathbf{w_{B}}$ are in a sufficiently close region $\mathbf{w}\in\sigma\left(\mathbf{w_{AB},w_{A}},\mathbf{w_{B}}\right)$. In addition to $\mathbf{w_{A}}$ and $\mathbf{w_{B}}$, we make the strong assumption that $\mathbf{w_{A}}$ and $\mathbf{w_{B}}$ have flat loss landscapes. By using coefficient comparisons, Eq.~(\ref{eq:loss_val}) can be derived. Sec.~\ref{sec:drive_loss_val} for the detailed derivation. This is an obvious consequence of strong assumptions. 

\subsection{Derivation of Eq.~(\ref{eq:loss_val})}
\label{sec:drive_loss_val}
Let us assume that $\mathbf{w_{AB}}$, $\mathbf{w_{A}}$ and $\mathbf{w_{B}}$ are close enough and loss landscapes are flat enough as $\mathbf{w}F_{A}\mathbf{w}=\mathbf{w}F_{B}\mathbf{w}\approx0$. We can show 
\begin{align}
L_{A}\left(\frac{1}{2}\left(\mathbf{w_{A}}+\mathbf{w_{B}}\right)\right)&=L_{A}\left(\mathbf{w_{A}}\right)+\frac{1}{2}\left(\frac{1}{2}\left(\mathbf{w_{A}}+\mathbf{w_{B}}\right)-\mathbf{w_{A}}\right)F_{A}\left(\frac{1}{2}\left(\mathbf{w_{A}}+\mathbf{w_{B}}\right)-\mathbf{w_{A}}\right)\nonumber\\&=L_{A}\left(\mathbf{w_{A}}\right)+\frac{1}{8}\left(\mathbf{w_{B}}-\mathbf{w_{A}}\right)F_{A}\left(\mathbf{w_{B}}-\mathbf{w_{A}}\right)\nonumber\\&=L_{A}\left(\mathbf{w_{A}}\right).
\end{align}
The same applies to $L_{B}\left(\frac{1}{2}\left(\mathbf{w_{A}}+\mathbf{w_{B}}\right)\right)$. By using this, it can be shown that
\begin{align}
L_{AB}\left(\frac{1}{2}\left(\mathbf{w_{A}}+\mathbf{w_{B}}\right)\right)&=\frac{1}{2}L_{A}\left(\frac{1}{2}\left(\mathbf{w_{A}}+\mathbf{w_{B}}\right)\right)+\frac{1}{2}L_{B}\left(\frac{1}{2}\left(\mathbf{w_{A}}+\mathbf{w_{B}}\right)\right)\nonumber\\&=\frac{1}{2}\left(L_{A}\left(\mathbf{w_{A}}\right)+L_{B}\left(\mathbf{w_{B}}\right)\right)\label{eq:approxymate}.
\end{align}
Then, as with Sec.~\ref{sec:drive_direction}, comparing the both sides of Eq.~(\ref{eq:data_linear}), we find that
\begin{align}
L_{AB}\left(\mathbf{w_{AB}}\right)&=\frac{1}{2}\left(L_{A}\left(\mathbf{w_{A}}\right)+L_{B}\left(\mathbf{w_{B}}\right)\right)-\mathbf{w_{AB}}F_{AB}\mathbf{w_{AB}}+\frac{1}{2}\left(\mathbf{w_{A}}F_{A}\mathbf{w_{A}}+\mathbf{w_{B}}F_{B}\mathbf{w_{B}}\right)\nonumber\\&=L_{AB}\left(\frac{1}{2}\left(\mathbf{w_{A}}+\mathbf{w_{B}}\right)\right)-\frac{1}{2}\mathbf{w_{AB}}\left(F_{A}+F_{B}\right)\mathbf{w_{AB}}+\frac{1}{2}\left(\mathbf{w_{A}}F_{A}\mathbf{w_{A}}+\mathbf{w_{B}}F_{B}\mathbf{w_{B}}\right)\nonumber\\&=L_{AB}\left(\frac{1}{2}\left(\mathbf{w_{A}}+\mathbf{w_{B}}\right)\right).
\end{align}
The second line of the transformation used Eq.~(\ref{eq:approxymate}). We can obtain $L_{AB}\left(\mathbf{w}_{AB}\right)=L_{AB}\left(\frac{1}{2}\left(\mathbf{w_{A}}+\mathbf{w_{B}}\right)\right)$.

\subsection{Extending the loss barrier of a single dataset to different datasets}\label{sec:lossbarrier}
This study clarifies the relationship between the loss barrier on a single dataset and the loss barrier on different datasets. The loss barrier in a single dataset is defined as 
\begin{align}
{\rm LB}=\max_{\lambda\in[0,1]}\left[\mathcal{L}\left(\mathbf{w_{M}}\right)-\left(1-\lambda\right)\mathcal{L}\left(\mathbf{w_{A}}\right)-\lambda\mathcal{L}\left(\mathbf{w_{B}}\right)\right],
\end{align}
where $\mathbf{w_{M}}=\left(1-\lambda\right)\mathbf{w_{A}}+\lambda \mathbf{w_{B}}$. In the STE on different datasets, $\lambda$ is fixed for the optimization target. Therefore, it is defined with respect to the fixed $\lambda$, as follows
\begin{align}
{\rm LB} & =\mathcal{L}_{AB}\left(\mathbf{w_{M}}\right)-\left(1-\lambda\right)\mathcal{L}_{AB}\left(\mathbf{w_{A}}\right)-\lambda\mathcal{L}_{AB}\left(\mathbf{w_{B}}\right)\nonumber\\
 & =\mathcal{L}_{AB}\left(\mathbf{w_{M}}\right)+{\rm const}\nonumber\\
 & =\frac{1}{2}\left(\mathcal{L}_{A}\left(\mathbf{w_{M}}\right)+\mathcal{L}_{B}\left(\mathbf{w_{M}}\right)\right)
\end{align}
The second line, $\mathcal{L}_{AB}\left(\pi\left(\mathbf{w_{B}}\right)\right)=\mathcal{L}_{AB}\left(\mathbf{w_{B}}\right)$, is considered constant because it remains unchanged under permutation. Therefore, it was demonstrated that the loss barrier used in model merging between different datasets is a natural extension of the loss barrier on a single dataset~\citep{ainsworth2023git}.

\section{Additional Experiments}
\subsection{Preliminary Experiments for OT Fusion}
\begin{table}[tb]
\caption{Test accuracy on MNIST (\%).}\label{tab:mnist_acc_ot}
\centering 
\begin{tabular}{@{}lcccc@{}}
\toprule
 & Model A & Model B & Model AB (Na\"ive) & Model AB (OT) \\ \midrule
$\lambda=0.5$, ${\rm sr}=0.1$ & 95.29 & 88.06 & 80.25 & 85.98 \\
$\lambda=0.5$, ${\rm sr}=0$ & 9.82 & 88.21 & 13.06 & 10.1 \\
$\lambda=0.2$, ${\rm sr}=0.1$ & 95.29 & 88.06 & 82.07 & 94.43 \\
$\lambda=0.2$, ${\rm sr}=0$ & 9.82 & 88.21 & 9.82 & 10.1 \\ \bottomrule
\end{tabular}
\end{table}
To evaluate the OT performance in the case of complete label separation, we conducted experiments varying the hyperparameters in Fig.~2 of the OT paper~\citep{singh2020model}. In Fig.~2 of the OT paper, Model A was trained using Label 4 and $10\%$ of labels other than 4, and Model B was trained using $100\%$ of data other than Label 4. Table~\ref{tab:mnist_acc_ot} shows the test accuracy on the MNIST data. The $\lambda$ is the coefficients of model combination, and ${\rm sr}$ is the percentage of non-specific labels used. For example, if we use $10\%$ of labels other than Label 4 and 4 for Model A, then ${\rm sr}=0.1$. Here, experiments were conducted with ${\rm sr}=0.0$ as the difficult task, rather than ${\rm sr}=0.1$ used in the paper. We also experimented with $\lambda=0.5$, which is the simple average of the model parameters, and $\lambda=0.2$, which is the best parameter in the paper. Since MNIST is a simple dataset, even with ${\rm sr}=0.1$, Model-A achieves $95\%$ accuracy, while with ${\rm sr}=0.0$, only $4$ labels are trained, which reduces the accuracy to $10\%$. For $\lambda=0.2$ and ${\rm sr}=0.1$, the accuracy of Model-AB (OT) almost reproduces the previous study values with $82\%$. When ${\rm sr}=0$, the test accuracy of Model AB (OT) is $10\%$, indicating that OT has difficulty model merging between completely separated labels. For ${\rm sr}=0.1$, Model A contains all the label information, so it is optimal to merge Model A more as $\lambda=0.2$. On the other hand, for ${\rm sr}=0$, Model A is completely independent of the label, so it is optimal to merge half of the model as $\lambda=0.5$ is considered to be the optimal case of mixing half and half as $\lambda=0.5$.

\subsection{Failed Idea: Fisher Connection}
\label{sec:fisher_connect}
\begin{figure*}
\centering
\begin{minipage}[t]{0.8\linewidth}
\includegraphics[width=0.49\linewidth]{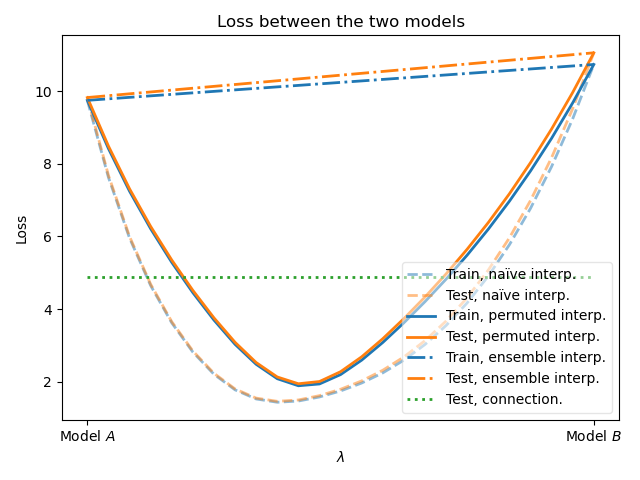}
\includegraphics[width=0.49\linewidth]{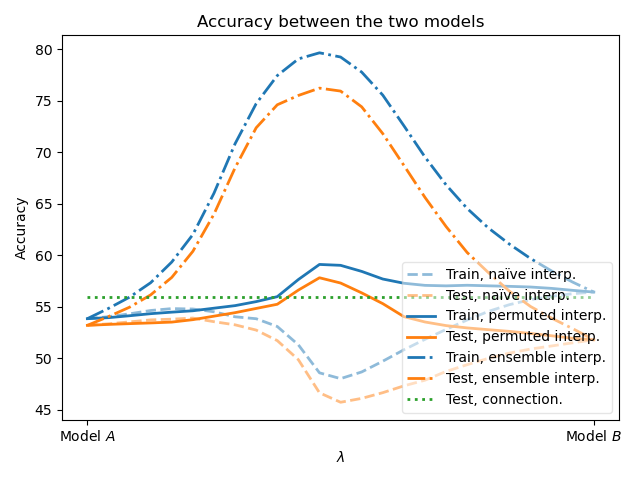}
\end{minipage}
\caption{Fisher strategy}\label{fig:fihser-connection}
\end{figure*}
We show the advantage of switching from the mean strategy to Fisher strategy for model merging after performing WM as Eq.~(\ref{eq:direction}). Figure~\ref{fig:fihser-connection} shows the loss and accuracy between MNIST and FMNIST. The green line represents the case of merging models using $\mathbf{w_{AB}}=\frac{1}{2}\left(F_{AB}\right)^{-1}\left(F_{A}\mathbf{w_{A}}+F_{B}\mathbf{w_{B}}\right)$ on $\lambda=0.5$. The Fisher strategy is less accurate than the average strategy. The Fisher strategy fails because it is numerically unstable due to the use of eigenvector directions with small eigenvalues of the Fisher information matrix around the flat solution.

\subsection{Failed Idea: Flat Weight Matching}
\label{sec:flat_weight_matching} 
\begin{figure*}
\centering
\begin{minipage}[t]{0.4\linewidth}
\includegraphics[width=1\linewidth]{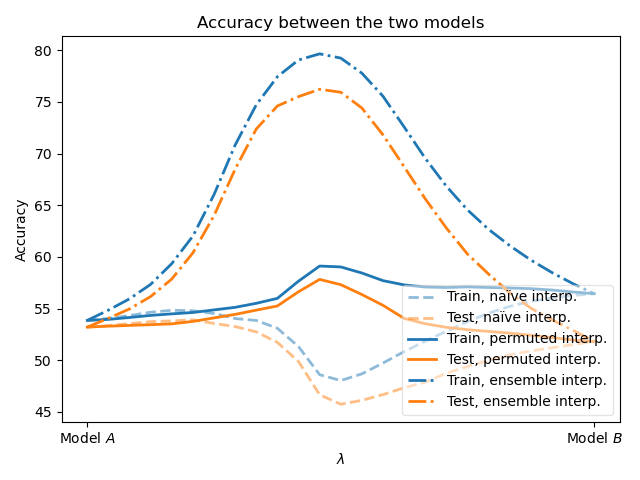}
\subcaption{WM}
\end{minipage}
\begin{minipage}[t]{0.4\linewidth}
\includegraphics[width=1\linewidth]{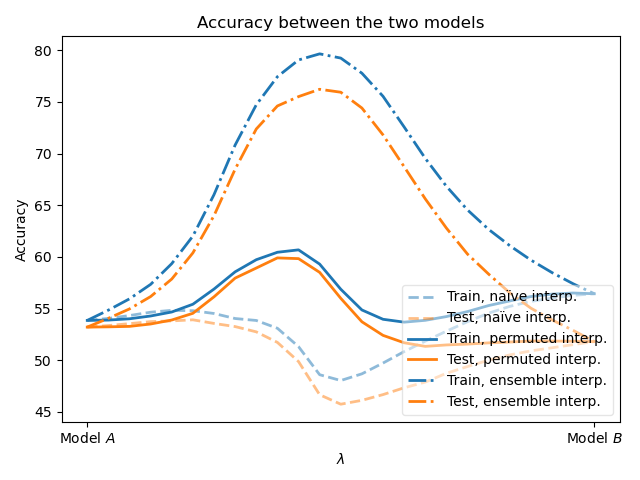}
\subcaption{FWM}
\end{minipage}
\caption{Acc with FWM on MNIST-FMNIST}\label{fig:flat-wm}
\end{figure*}
Section~\ref{sec:merge-condition} shows that the flatness of the loss landscape is important for successful model merging. Thus, we propose Flat Weight Matching~(FWM), which adds to the cost function of WM the regularization that not only the L2 distance is close but also the loss landscape is flat as
\begin{align}
\argmin_{\pi}\left[\beta\left\Vert {\rm vec}\left(\mathbf{w_{A}}\right)-{\rm vec}\left(\pi\left(\mathbf{w_{B}}\right)\right)\right\Vert ^{2}+\left(1-\beta\right)\left({\rm vec}\left(\mathbf{w_{A}}\right)-{\rm vec}\left(\pi\left(\mathbf{w_{B}}\right)\right)\right)\left.\frac{\partial\mathcal{L}_{B}}{\partial\mathbf{w}}\right|_{\mathbf{w}=\pi\left(\mathbf{w_{B}}\right)}\right],
\end{align}
where $\beta$ is the hyperparameter and the second term is a regularization that promotes flattening the loss landscape. The loss landscape of Model A is flat because it is optimized by SGD. Since the gradient of the permuted weight can be computed from a permutation of the pre-computed gradient, FWM has the advantage that the gradient does not need to be recomputed. In our experiments, we used $\beta=0.01$, which performed the best among $\beta\in\left[0,1\right]$. Fig.~\ref{fig:flat-wm} compares WM and FWM. Improvement with FWM was marginal. This is because regularization using local gradients on the optimal weights could not sufficiently promote flatness as shown in Fig.~\ref{fig:acc-losslandscape}.

\subsection{Failed Idea: WM and Fine-tuning with the Condensed Dataset.}
To train on data AB without using the full dataset, we propose fine-tuning on a surrogate Dataset AB, starting with well-chosen initial weights. Here, we create good initial weights by using WM to minimize the L2 distance of the weights. Table~\ref{tab:wm-ft} compares Data Cond across multiple datasets, showing a mix of improvements and deteriorations, thus lacking consistency. As shown in Table~\ref{tab:summary}, training from scratch using Data Cond results in higher accuracy than Coreset, which led us to employ Data Cond.
\begin{table}[H]
\caption{Test accuracy on $D_{AB}$ using WM as initial weights and fine-tuning with the condensed dataset.}\label{tab:wm-ft}
\centering
\small
\begin{tabular}{lcccc}
\hline
                  & MNIST-RMNIST         & USPS-MNIST & SPLIT-CIFAR10 & MNIST-FMNIST \\ \hline
Data Cond          & 74.47        & 82.49      & 38.71         & 73.28        \\                
WM+FT (Data Cond) & 69.60                & 91.90      & 52.60         & 81.26        \\ \hline
\end{tabular}
\end{table}

\subsection{Model Merging on Single Dataset}
\label{sec:same-data}
\begin{figure*}
\centering
\begin{minipage}[t]{0.3\linewidth}
\includegraphics[width=1\linewidth]{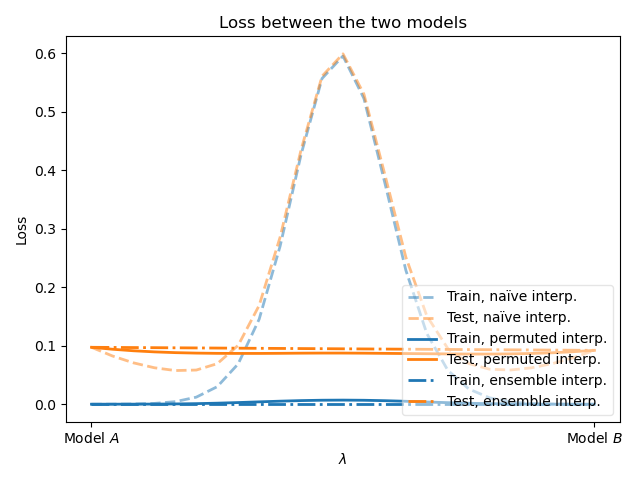}
\subcaption{MNIST}
\end{minipage}
\begin{minipage}[t]{0.3\linewidth}
\includegraphics[width=1\linewidth]{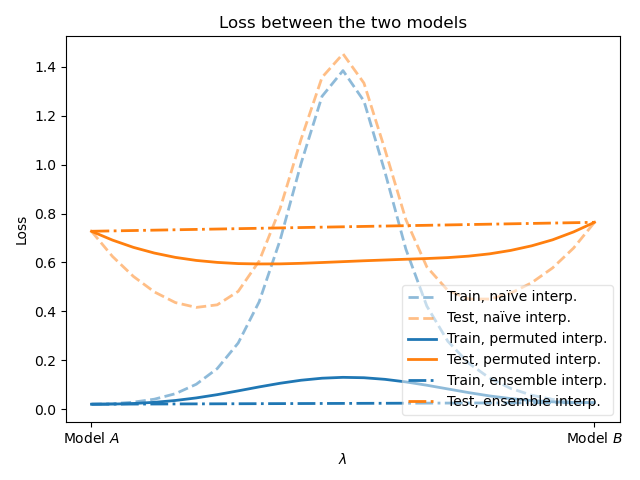}
\subcaption{FMNIST}
\end{minipage}
\begin{minipage}[t]{0.3\linewidth}
\includegraphics[width=1\linewidth]{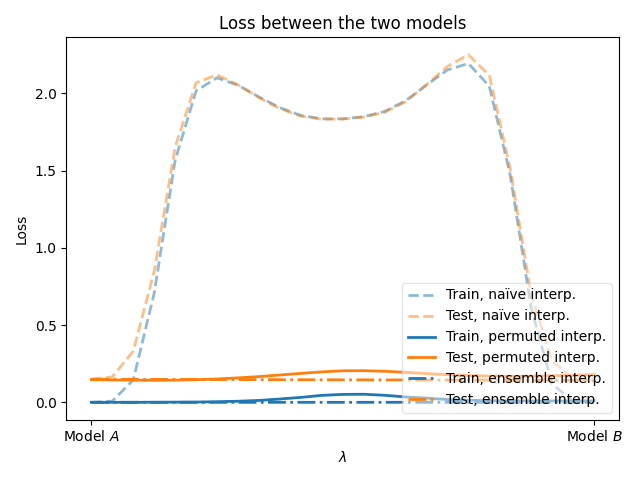}
\subcaption{SPLIT-CIFAR10-A with width multiplier 4}
\end{minipage}
\begin{minipage}[t]{0.3\linewidth}
\includegraphics[width=1\linewidth]{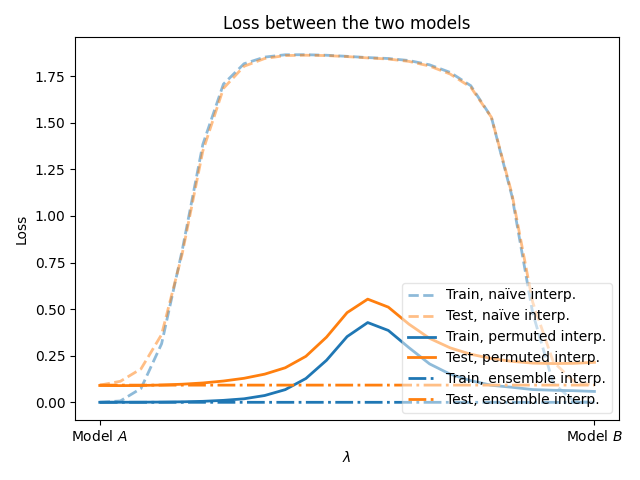}
\subcaption{SPLIT-CIFAR10-B with width multiplier 4}
\end{minipage}
\begin{minipage}[t]{0.3\linewidth}
\includegraphics[width=1\linewidth]{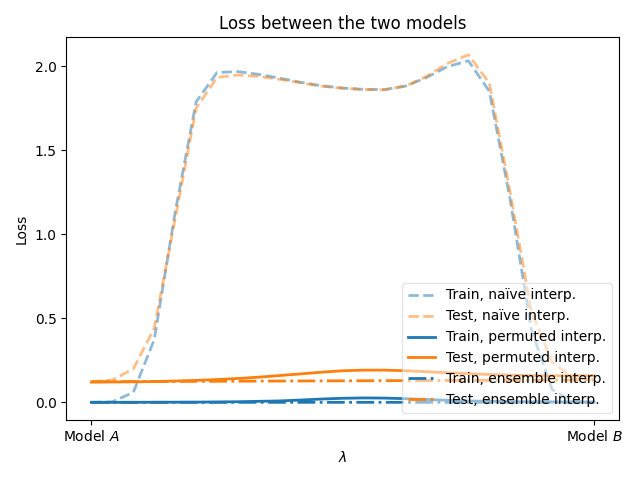}
\subcaption{SPLIT-CIFAR10-A with width multiplier 16}
\end{minipage}
\begin{minipage}[t]{0.3\linewidth}
\includegraphics[width=1\linewidth]{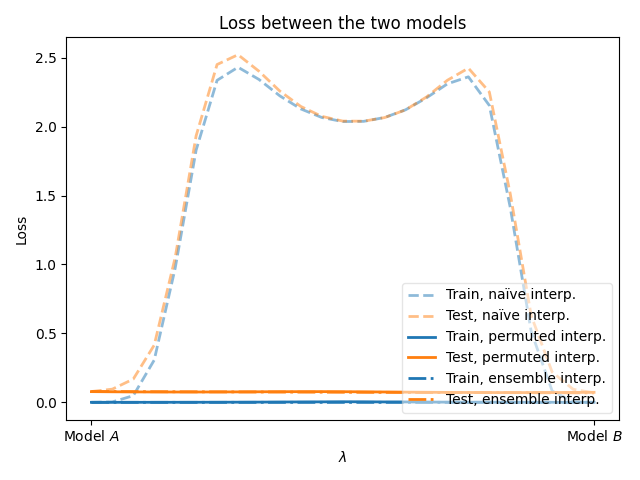}
\subcaption{SPLIT-CIFAR10-B with width multiplier 16}
\end{minipage}
\caption{Loss of the merged model on single datasets.}
\label{fig:same-data-single-loss}
\end{figure*}

\begin{figure*}
\centering
\begin{minipage}[t]{0.3\linewidth}
\includegraphics[width=1\linewidth]{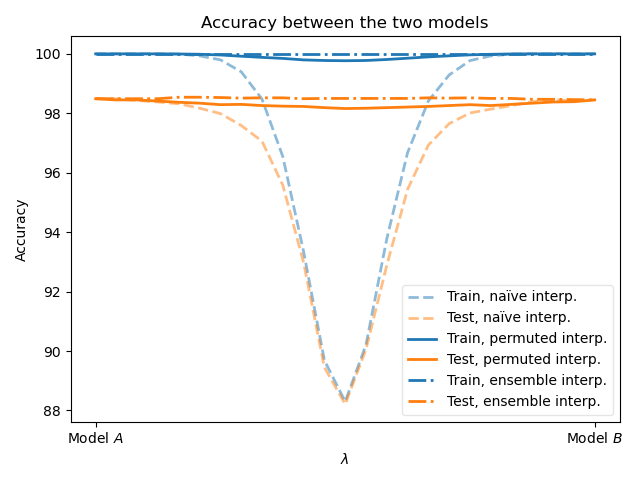}
\subcaption{MNIST}
\end{minipage}
\begin{minipage}[t]{0.3\linewidth}
\includegraphics[width=1\linewidth]{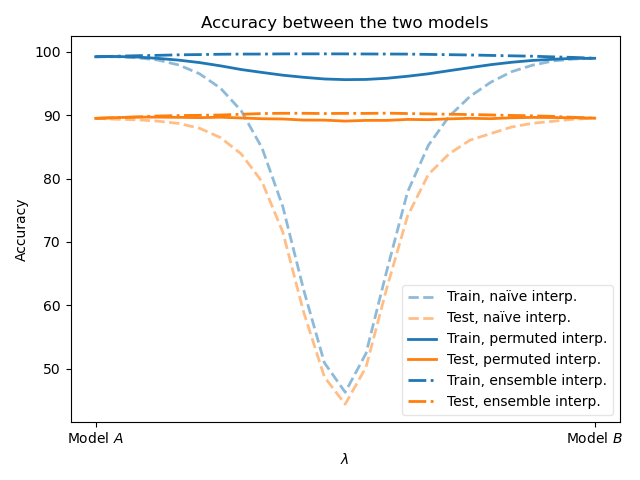}
\subcaption{FMNIST}
\end{minipage}
\begin{minipage}[t]{0.3\linewidth}
\includegraphics[width=1\linewidth]{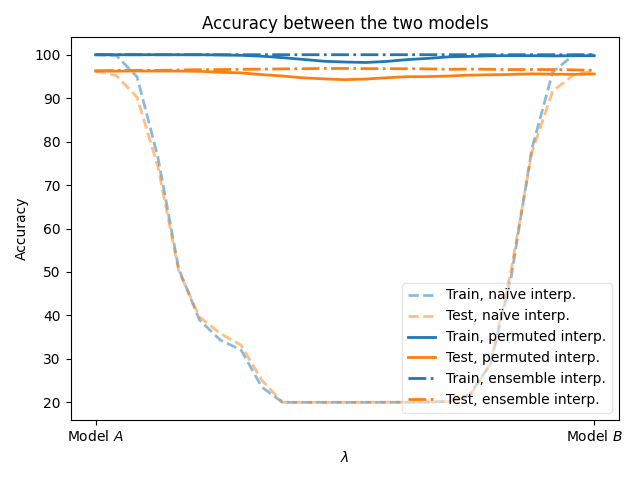}
\subcaption{SPLIT-CIFAR10-A with width multiplier 4}
\end{minipage}
\begin{minipage}[t]{0.3\linewidth}
\includegraphics[width=1\linewidth]{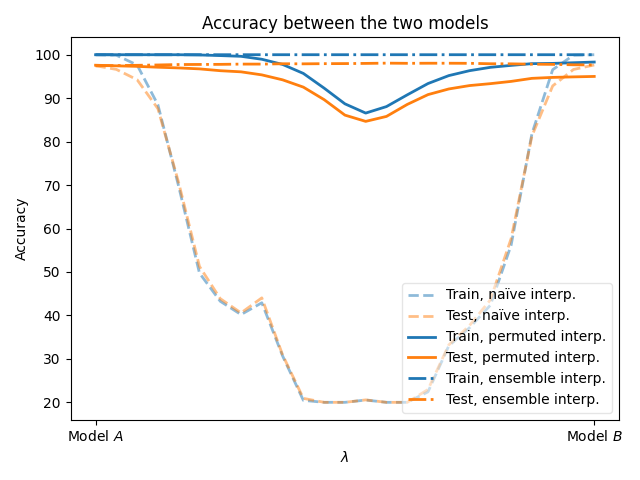}
\subcaption{SPLIT-CIFAR10-B with width multiplier 4}
\end{minipage}
\begin{minipage}[t]{0.3\linewidth}
\includegraphics[width=1\linewidth]{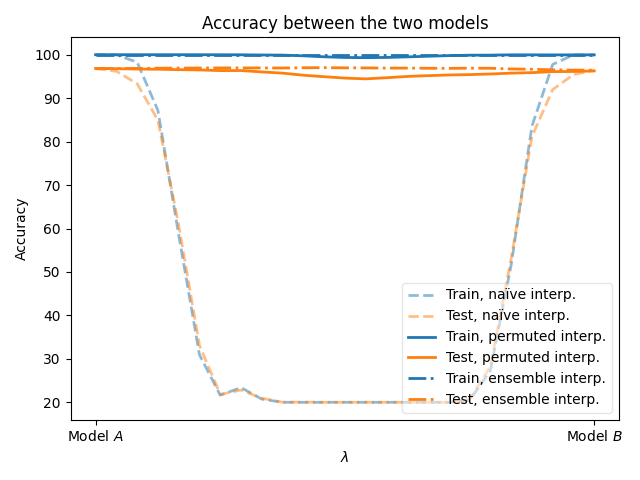}
\subcaption{SPLIT-CIFAR10-A with width multiplier 16}
\end{minipage}
\begin{minipage}[t]{0.3\linewidth}
\includegraphics[width=1\linewidth]{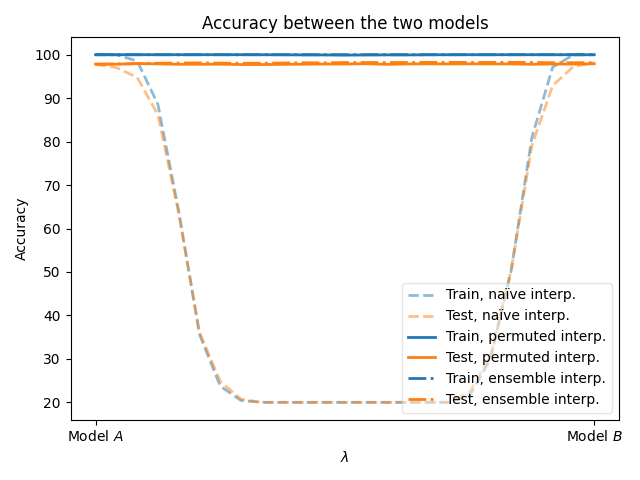}
\subcaption{SPLIT-CIFAR10-B with width multiplier 16}
\end{minipage}
\caption{Acc of the merged model on single datasets.}
\label{fig:same-data-single-acc}
\end{figure*}
This subsection presents preliminary experiments merging models by using WM on a single dataset. Figures~\ref{fig:same-data-single-loss} and ~\ref{fig:same-data-single-acc} show loss and accuracy on MNIST, FMNIST, and SPLIT-CIFAR10, where SPLIT-CIFAR10 was used with ResNet20 widths of 4 and 16 times. MNIST and FMNIST successfully merge models with high accuracy in each case. SPLIT-CIFAR10 allowed for merging models with high accuracy by increasing the width by a factor of 16. For model merging between different datasets, we used merging models that can be merged on a single dataset.

\subsection{Effect of Model Width}
This section discusses the impact of model width on the success of model merging. It is known that LMC on a single dataset is more valid the larger the model width is~\citep{ainsworth2023git}. Since models are difficult to merge on different datasets if they are not merging well on a single dataset, the larger the model width on different datasets, the higher the accuracy after merging tends to be.

\subsection{Model Merging with Coreset selection}\label{sec:modelmerge-coreset}
\begin{figure*}
\centering
\begin{minipage}[t]{0.32\linewidth}
\includegraphics[width=\linewidth]{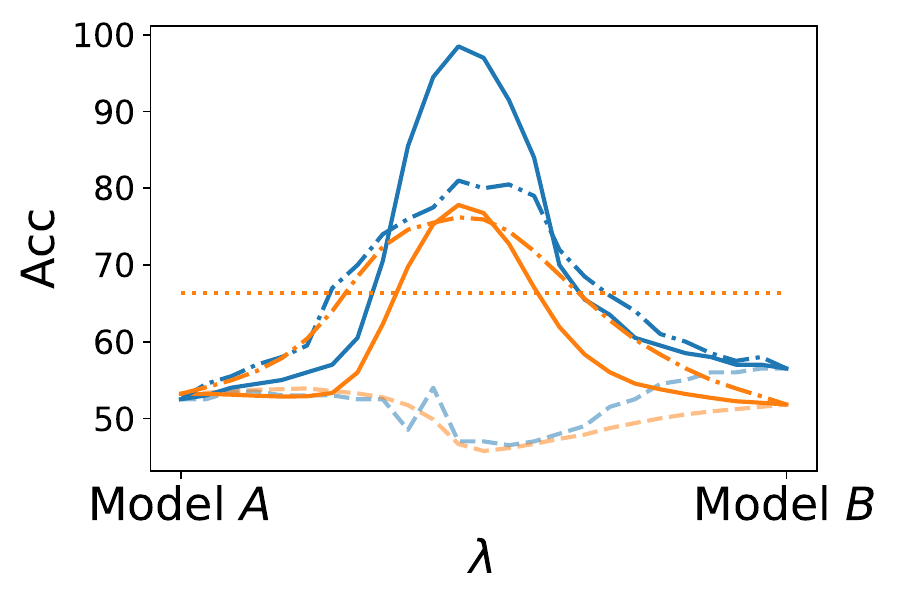}
\subcaption{Random}
\end{minipage}
\begin{minipage}[t]{0.32\linewidth}
\includegraphics[width=\linewidth]{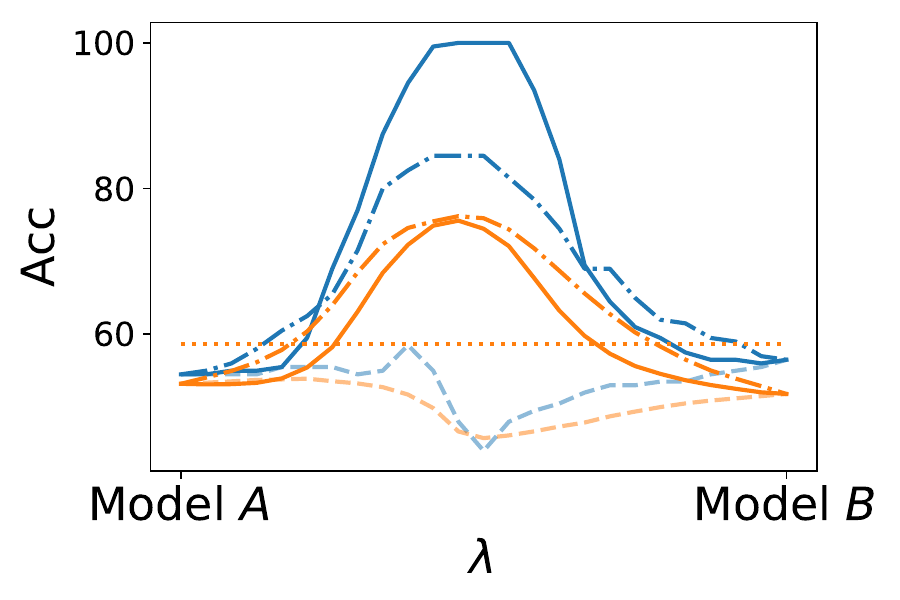}
\subcaption{GC}
\end{minipage}
\begin{minipage}[t]{0.32\linewidth}
\includegraphics[width=\linewidth]{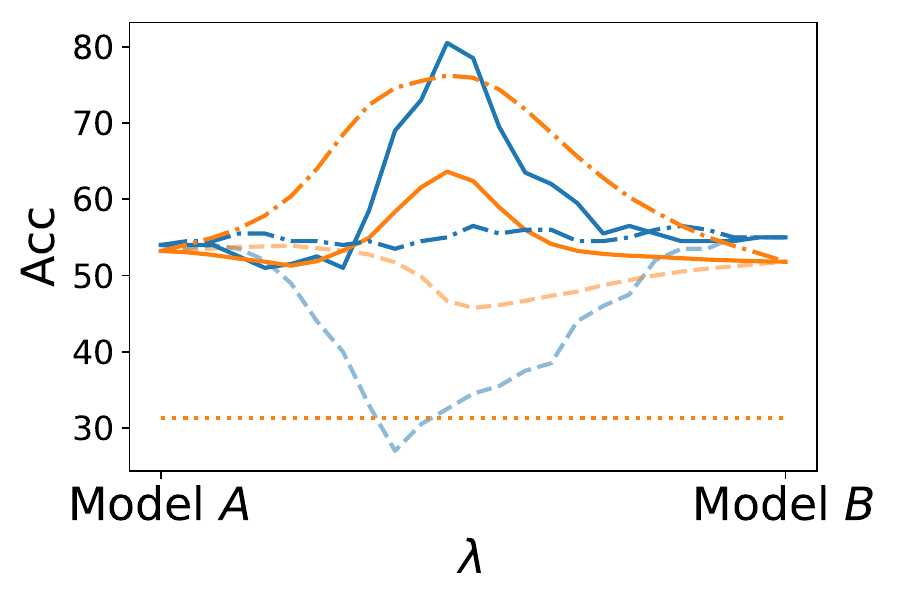}
\subcaption{Margin}
\end{minipage}
\begin{minipage}[t]{0.32\linewidth}
\includegraphics[width=\linewidth]{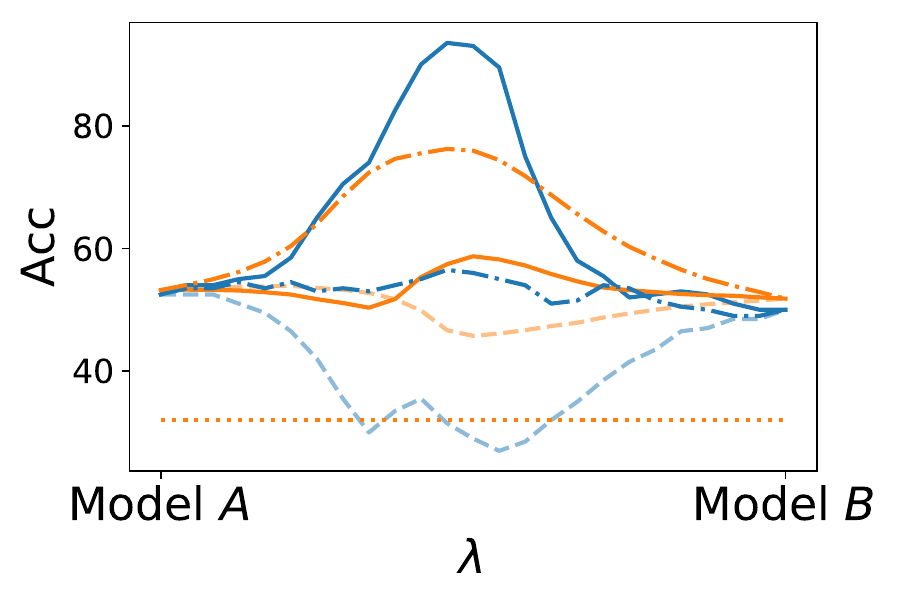}
\subcaption{Glister}
\end{minipage}
\begin{minipage}[t]{0.32\linewidth}
\includegraphics[width=\linewidth]{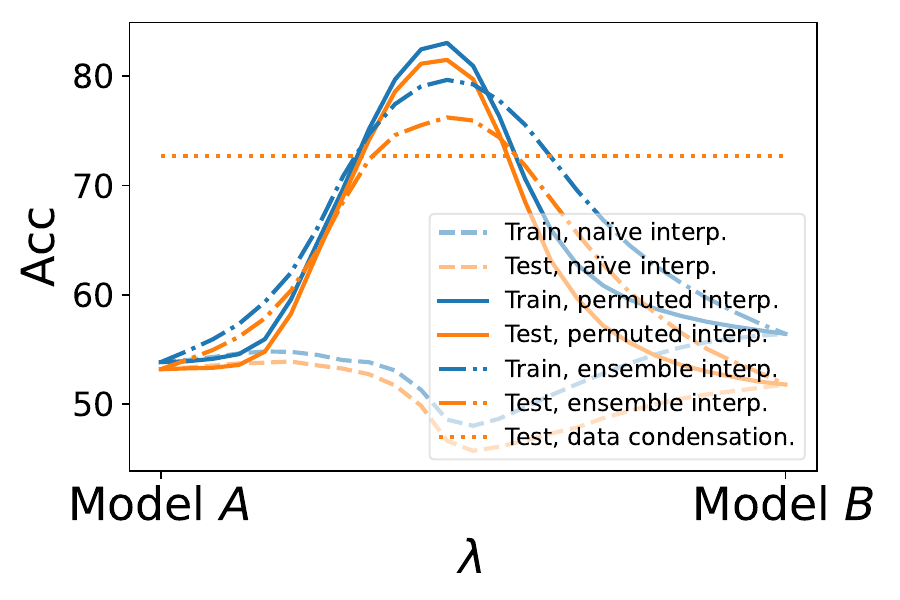}
\subcaption{Data Cond-10 (Reference)}
\end{minipage}
\caption{\textbf{Model Merge with Coreset Selection on MNIST-FMNIST.} These figures show the test accuracy of merging models using condensed datasets created with coreset selection.}
\label{fig:corset-depend}
\end{figure*}

Figures~\ref{fig:corset-depend} present the STE outcomes using various coreset selection methods, all of which demonstrate significantly higher accuracy than Na\"ive. Notably, some methods, including random sampling, outperform ensemble. This is an important result for achieving easy-to-use model merging. We compared four representative methods of different properties. For these experiments, we used DeepCore~\citep{guo2022deepcore} which is a comprehensive library for coreset selection. We also conducted ten trials with different random seeds for random sampling to assess its dependence on random number generation. The results indicate minimal influence from random variation

\subsection{Model Merging for Diverse Training Conditions: Variation in Loss Function and Optimizers}
\begin{table}[H]
\caption{\textbf{Comparison of Merged Model Outcomes using Different Loss Functions and Optimizers.} Model A was trained using softmax loss and SGD with a learning rate of $0.1$, without weight decay. Conversely, Model B was trained and subsequently merged with STE (full) using various conditions: softmax with label-smoothing regularization, the Adam optimizer, SGD with weight decay, and SGD with a learning rate of $0.05$.}\label{tab:losses-optimizers}
\centering
\begin{tabular}{@{}ccccc@{}}
\toprule
Difference    & Label Smoothing & Adam  & Weight Decay & Learning Rate \\ \midrule
Test Acc (\%) & 93.87           & 92.61 & 93.72        & 93.55         \\ \bottomrule
\end{tabular}
\end{table}
Additional experiments of merging models between different losses and optimizers were performed to clarify the limitations of the experiment. As an experimental setup, Model A was trained with softmax loss SGD with a learning rate (lr) of $0.1$ without weight decay, and Model B was trained and merged models with STE full in four cases: (1) softmax with label-smoothing regularization with $\epsilon=0.1$~\citep{szegedy2016rethinking}, (2) Adam with ${\rm lr}=0.0005$, (3) SGD with weight decay, and (4) SGD with ${\rm lr}=0.05$. The Test Acc on $D_{AB}$ of the merging models is shown in Table~\ref{tab:losses-optimizers}. In all cases, the merging models achieved an accuracy that outperformed the ensemble.

\subsection{Model Merging with varying mixing ratio $\alpha$ of datasets}
Focusing on $\alpha=0.5$ might seem overly idealized for real-world scenarios. To address this, we have conducted additional experiments varying the $\alpha$ value across a range of scenarios, including more unbalanced mixing ratios (e.g., $\alpha=0.2$ and $\alpha=0.8$) using the MNIST-RMNIST90 datasets. 
\begin{table}[h]
\centering
\begin{tabular}{@{}cccccccc@{}}
\toprule
$\alpha$ & 0 & 0.2 & 0.4 & 0.5 & 0.6 & 0.8 & 1.0 \\ \midrule
Test Acc (\%) & 98.26 & 92.58 & 92.32 & 91.55 & 91.42 & 92.43 & 98.45 \\ \bottomrule
\end{tabular}
\caption{Test Accuracy on MNIST-RMNIST with varying $\alpha$. Here, $\alpha=0.5$ means that half of the data was sampled from each dataset, but in the main paper, we used all data points to create a balanced mixed dataset.}
\label{tab:alpha_mnist_rmnist}
\end{table}

Interestingly, we found that the model merging method remains effective across different values of $\alpha$, and that $\alpha=0.5$ actually represents the most challenging scenario. This is likely because when $\alpha$ is closer to $0$ or $1$, the loss can be reduced more easily by relying more heavily on the weights from either Model A or Model B. These results suggest that our model merging approach is robust across various mixing ratios, and that $\alpha=0.5$ provides a more rigorous test of the method’s capabilities.

\subsection{Model Merging on SPLIT-CIFAR100}
\begin{table}[h]
\centering
\begin{tabular}{lccc}
\toprule
Method & STE (Full) & STE (Coreset) & ZipIt! (Full) \\ \midrule
Test Acc (\%) & 62.94 & 60.52 & 27.9 \\ \bottomrule
\end{tabular}
\caption{Test Accuracy on SPLIT-CIFAR100 for different methods. Here, we SPLIT-CIFAR100 into two sets of 50 classes each for training and then performed model merging. Our experimental setup is the same as ZipIt's CIFAR100 (50+50). STE (Coreset) utilizes samples of 50 images per class. }
\label{tab:split_cifar100}
\end{table}

\subsection{Loss Landscape}\label{sec:losslandscape}
\begin{figure*}
\centering
\begin{minipage}[t]{0.48\linewidth}
\includegraphics[width=0.43\linewidth]{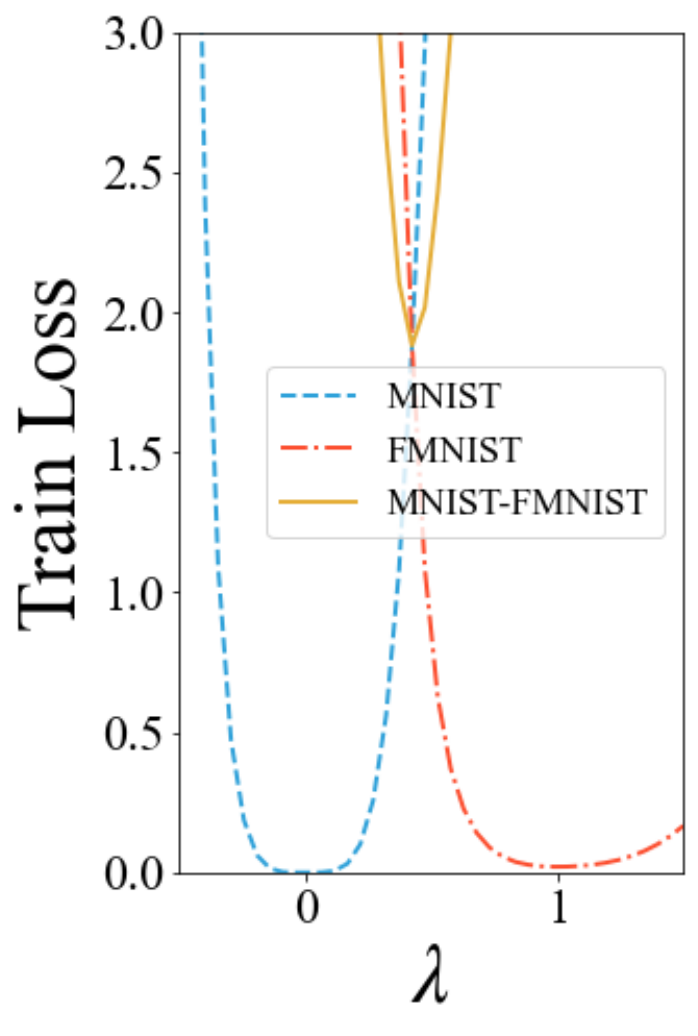}
\includegraphics[width=0.43\linewidth]{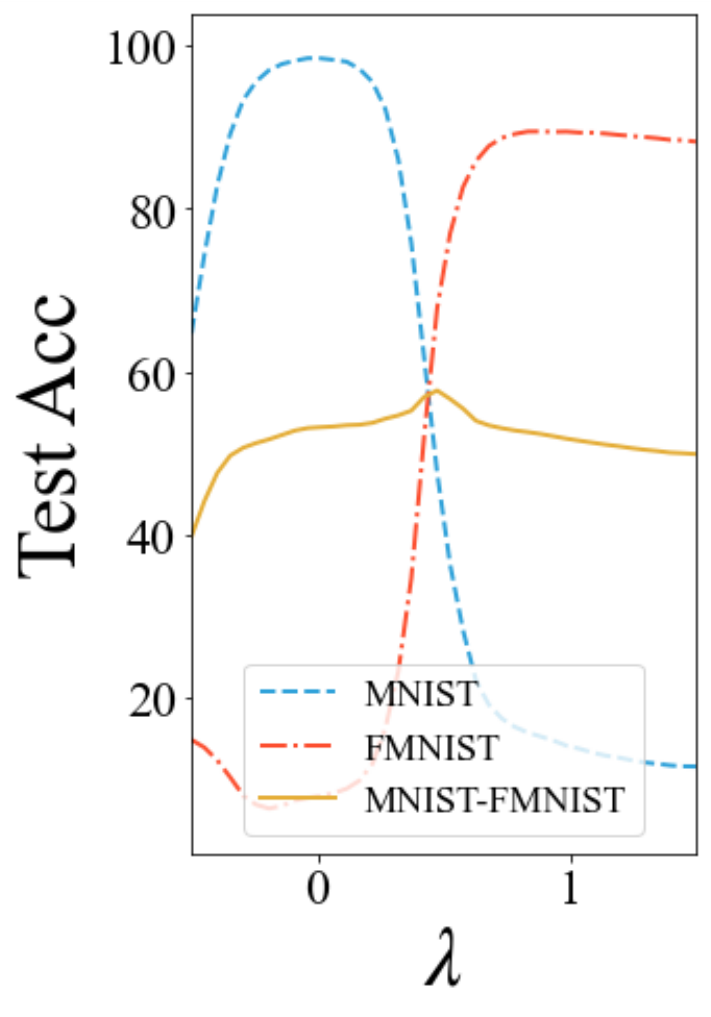}
\subcaption{WM}\label{fig:landscape-wm}
\end{minipage}
\begin{minipage}[t]{0.48\linewidth}
\includegraphics[width=0.43\linewidth]{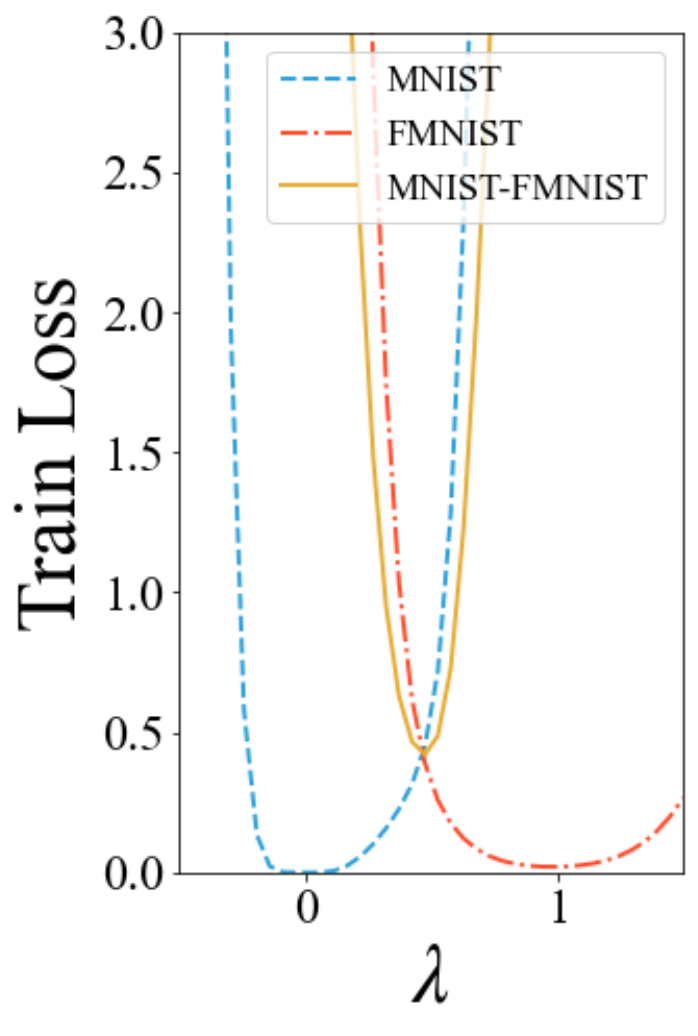}
\includegraphics[width=0.43\linewidth]{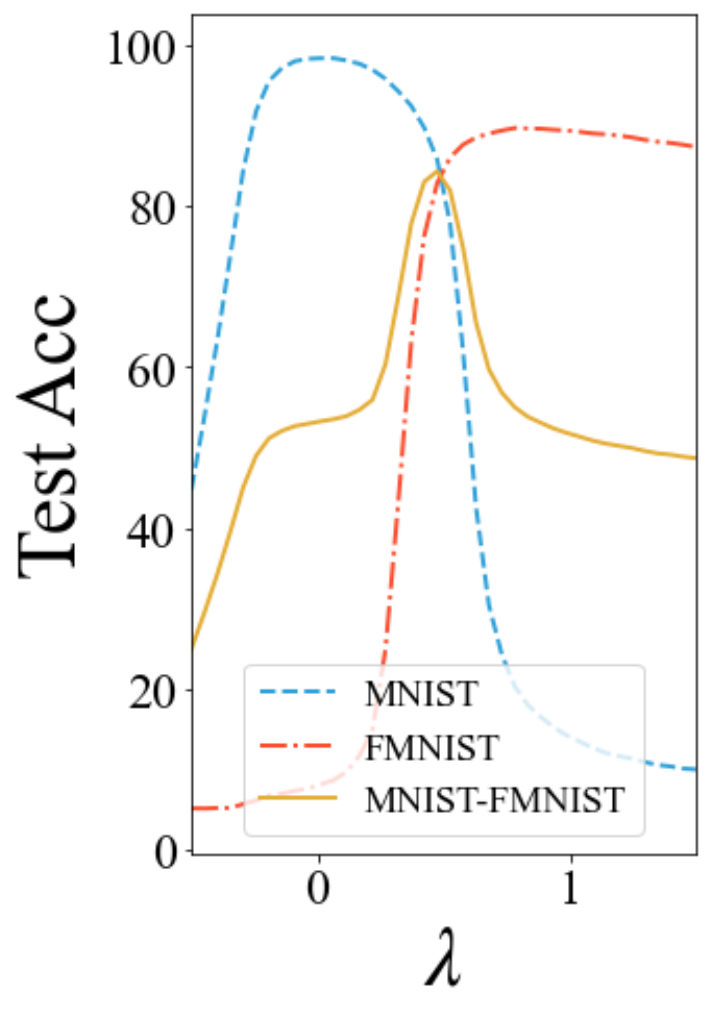}
\subcaption{STE}\label{fig:landscape-ste}
\end{minipage}
\caption{Training loss and test accuracy on MNIST-FMNIST. Line differences represent the different datasets to be evaluated, with $\lambda=0$ representing the optimal weights for MNIST and $\lambda=1$ representing the optimal weights for FMNIST.}
\label{fig:acc-losslandscape}
\end{figure*}

We investigate the loss landscape to clarify the differences in the performance of merging models in WM and STE. The investigation uses MNIST-FMNIST, model merging between completely different datasets. Figure~\ref{fig:acc-losslandscape} shows the loss landscape corresponding to Eq.~(\ref{eq:star-op}) on the MNIST-FMNIST, with $\lambda=0$ representing the optimal weights for MNIST and $\lambda=1$ representing the optimal weights for FMNIST. Figure~\ref{fig:acc-losslandscape} shows that for both WM and STE, $\mathbf{w_{AB}}$ and $\mathbf{w_{A}}$ and $\mathbf{w_{B}}$ are in regions close enough that the loss landscape can be considered a convex function. Comparing WM and STE shows that STE has a smaller loss barrier. This is because STE optimizes the direct reduction of the loss barrier, which leads to flatter loss landscapes.

\subsection{Overlap of Critical Weights}
\label{sec:overlap}
\begin{figure*}
\centering
\begin{minipage}[t]{0.25\linewidth}
\includegraphics[width=1\linewidth]{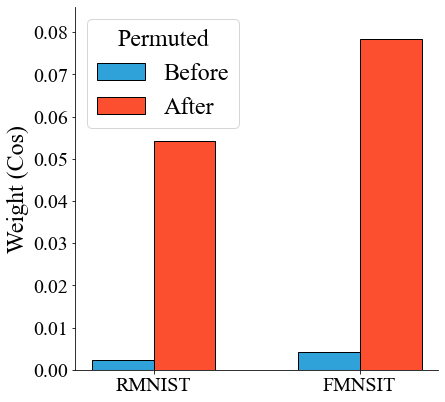}
\subcaption{Weight Vector}\label{fig:weight-cos}
\end{minipage}
\begin{minipage}[t]{0.25\linewidth}\centering
\includegraphics[width=1\linewidth]{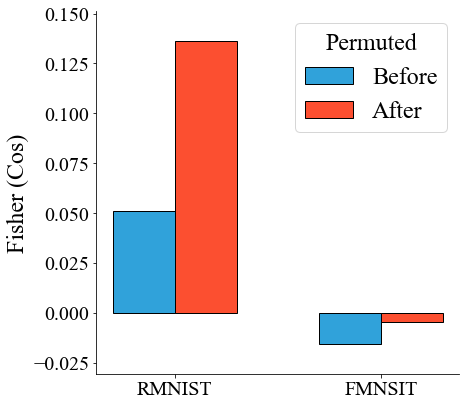}
\subcaption{Importance Vector}\label{fig:fisher-cos}
\end{minipage}
\caption{Overlap of weights and importance vector of weights between before and after permutations on MNIST-RMNIST and MNIST-FMNIST. Before and after permutation of RMNIST and FMNIST cos distance to MNIST.}
\label{fig:all-cos}
\end{figure*}
To reveal the properties of weight permutations, we measure the importance vector as the diagonal components of the Fisher information matrix and evaluate the overlap between Models A and B. Existing methods for learning multiple datasets, such as continuous learning, embed knowledge of each dataset into a single network. These embeddings avoid overlapping weights that are important for the inference of each dataset~\citep{kirkpatrick2017overcoming,fernando2017pathnet}. In other words, different forward propagation paths are used in inference for different datasets. Our definition of the importance vector is a one-dimensional arrangement of the diagonal components of the Fisher information matrix on mixed datasets, as in Kirkpatrick et al.~\citep{kirkpatrick2017overcoming}. The overlap of importance vectors is defined by the cos distance between the importance vectors of Models A and B. Figure~\ref{fig:fisher-cos} shows the results in MNIST-RMNIST and MNIST-FMNIST by using STE. They are the case of model merging with and without shared labels. The permutation for model merging shows that the overlap in the importance of the weights is not reduced. This is a different result from existing studies and does not indicate that different forward propagation paths are used for different datasets. Furthermore, Fig.~\ref{fig:weight-cos} shows the cos distance of the weights between Models A and B before and after the permutation. The permutation by STE also increased the cos distance of the weights as in the case of WM. This indicates that STE also reduces the L2 distance of weights between models.


\subsection{Evaluating the Effects of Merging Models Trained on Different Datasets}
\begin{table}[H]
\centering
\caption{\textbf{Evaluating the Effects of Merging Models Trained on Different Datasets.} The table shows the results of model merging using WM between a model trained on MNIST and models trained on other datasets. L2 (init) and L2 represent the L2 distance between the weights of Models A and B before and after applying WM, respectively.}
\begin{tabular}{@{}lccccc@{}}
\toprule
\multicolumn{1}{c}{Dataset} & \multicolumn{1}{c}{L2 (init)} & \multicolumn{1}{c}{L2} & \multicolumn{1}{c}{Barrier} & \multicolumn{1}{c}{FAcc} & \multicolumn{1}{c}{Acc~(WM)} \\ \midrule
FMNIST & $7.20 \times 10^{-5}$ & $6.70 \times 10^{-5}$ & 2.468 & 11.04 & 58.53 \\ 
RMNIST-$90^\circ$ & $4.61 \times 10^{-5}$ & $4.03 \times 10^{-5}$ & 1.280 & 14.79 & 70.86 \\ 
RMNIST-$75^\circ$ & $4.66 \times 10^{-5}$ & $4.06 \times 10^{-5}$ & 1.098 & 17.05 & 74.19 \\ 
RMNIST-$60^\circ$ & $4.68 \times 10^{-5}$ & $4.07 \times 10^{-5}$ & 0.804 & 26.27 & 80.33 \\ 
USPS & $9.69 \times 10^{-6}$ & $9.69 \times 10^{-6}$ & 0.450 & 42.83 & 85.19 \\ 
RMNIST-$45^\circ$ & $4.67 \times 10^{-5}$ & $4.05 \times 10^{-5}$ & 0.390 & 45.95 & 89.67 \\ 
RMNIST-$30^\circ$ & $4.65 \times 10^{-5}$ & $4.00 \times 10^{-5}$ & 0.217 & 75.26 & 94.69 \\ 
RMNIST-$15^\circ$ & $4.65 \times 10^{-5}$ & $3.96 \times 10^{-5}$ & 0.062 & 95.22 & 97.53 \\ 
RMNIST-$0^\circ$ & $4.60 \times 10^{-5}$ & $3.84 \times 10^{-5}$ & 0.008 & 98.47 & 98.49 \\ 
\bottomrule
\end{tabular}
\end{table}

\subsection{Model Merging between Multiple Datasets}
\begin{figure}[H]
\centering
\begin{minipage}{.6\textwidth}
  \centering
  \captionof{table}{\textbf{Accuracy Degradation in Model Merging between Three Datasets.} The parentheses indicate a stepwise model merging process. For MNIST+RMNIST+FMNIST, the average of three weights is used, where RMNIST and FMNIST are permuted to be compatible for merging with MNIST. Only in the ``Tuned'' case is $\lambda$ explored, while in all other cases, $\lambda$ is set to $0.5$.}\label{tab:three_model_acc}
  \begin{tabular}{@{}ll@{}}
    \toprule
    Model                       & Test Acc \\ \midrule
    MNIST                       & 53.18    \\
    RMNIST                      & 39.7     \\
    FMNIST                      & 42.43    \\ \midrule
    MNIST+RMNIST                & \bf{65.66}    \\
    MNIST+FMNIST                & 60.00    \\ \midrule
    (MNIST+RMNIST)+FMNIST       & 47.72    \\
    MNIST+(RMNIST+FMNIST)       & 53.19    \\
    MNIST+RMNIST+FMNIST         & 53.18    \\
    MNIST+RMNIST+FMNIST (Tuned) & \bf{65.66}    \\ \bottomrule
  \end{tabular}
\end{minipage}%
\begin{minipage}{.38\textwidth}
  \centering
  \includegraphics[width=\linewidth]{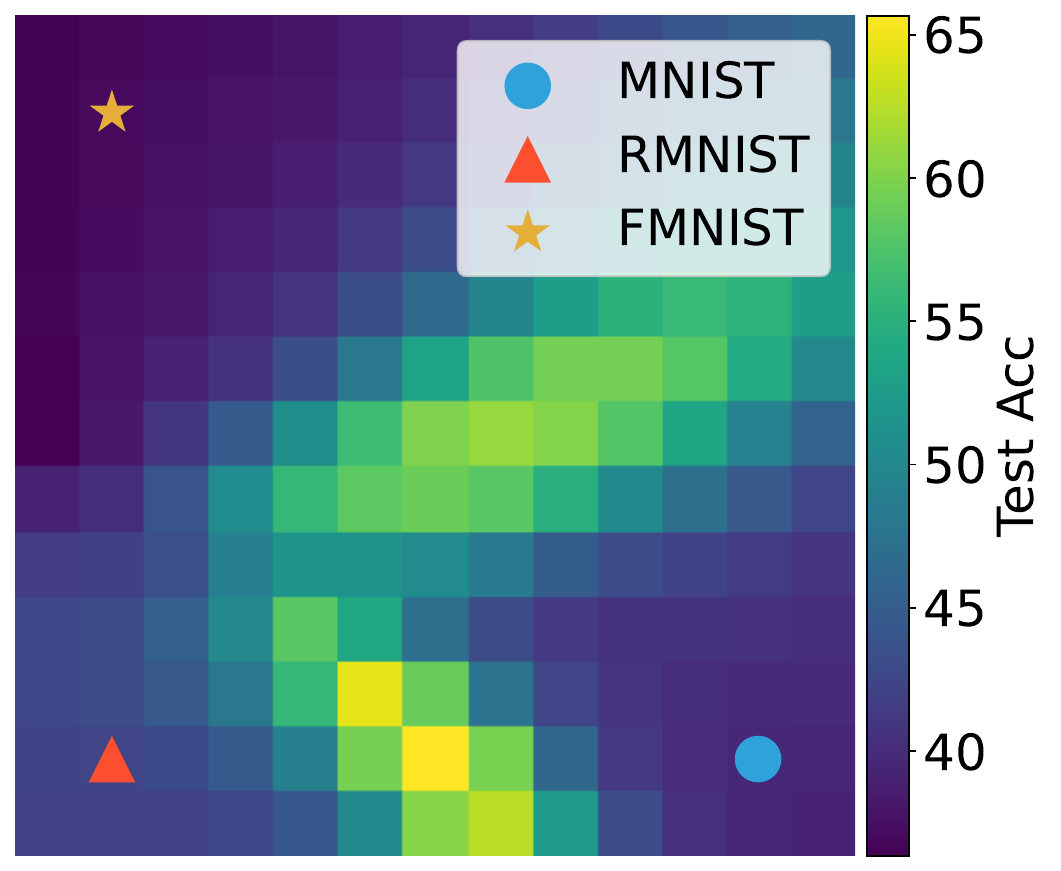}
  \captionof{figure}{\textbf{Test accuracy landscape  on MNIST-RMNIST-FMNIST around optimal weights.}}
  \label{fig:three_accuracy_landscape}
\end{minipage}
\end{figure}
In this section, we evaluate the merging of two or more models, focusing on the combination of MNIST, RMNIST, and FMNIST. Using MNIST as the anchor model, the average of three weights is used for the merging, where RMNIST and FMNIST are permuted to be compatible for merging with MNIST. Table~\ref{tab:three_model_acc} presents the test accuracy on mixed datasets of MNIST-RMNIST-FMNIST. The fact that the accuracy does not differ significantly from that of MNIST+RMNIST indicates that the merging of three models is not effectively achieved. Furthermore, although we attempted a stepwise merging of the three models, this approach did not result in improved outcomes. To elucidate the cause of this, Figure~\ref{fig:three_accuracy_landscape} illustrates the accuracy landscape. It is evident that only the combinations between MNIST and RMNIST, and MNIST and FMNIST, which explicitly reduced the loss barrier through STE, demonstrate high accuracy.
\end{document}